\documentclass[sigconf,nonacm]{acmart}
\AtBeginDocument{%
  }

\setcopyright{acmlicensed}
\copyrightyear{2018}
\acmYear{2018}
\acmDOI{XXXXXXX.XXXXXXX}
\acmConference[Conference acronym 'XX]{Make sure to enter the correct
  conference title from your rights confirmation email}{June 03--05,
  2018}{Woodstock, NY}
\acmISBN{978-1-4503-XXXX-X/2018/06}

\usepackage{subcaption}
\usepackage{graphicx}
\usepackage[figuresright]{rotating}
\usepackage{multirow} 
\usepackage{tabularx} % for 'tabularx' env. and 'X' col. type
\usepackage{color}
\usepackage{colortbl}
\usepackage{enumitem} 
\usepackage[linesnumbered,ruled,vlined]{algorithm2e}
\usepackage{xcolor}

\definecolor{myblue}{RGB}{44, 115, 183} 
\definecolor{myorange}{RGB}{219, 87, 44}
\usepackage{amsthm}

\begin{document}

\title{Just for FUNS: LLM-Guided Spatio-Temporal Graph Node Generation for Forecasting Unobserved Node States}

\author{Shuhao Li}
\affiliation{%
  \institution{ Fudan University}
  \city{Shanghai}
  \country{China}}
  \affiliation{%
  \institution{Nanyang Technological University}
   \country{Singapore}}
\email{shli23@m.fudan.edu.cn}

\author{Weidong Yang}
\affiliation{%
  \institution{Fudan University}
  \city{Shanghai}
  \country{China}}
\affiliation{%
  \institution{Zhuhai Fudan Innovation Research Institute }
  \city{Zhuhai}
  \country{China}}

\author{Changan Liu}
\affiliation{
  \institution{College of Computing and Data Science\\Nanyang Technological University}
  \country{Singapore}
}

\author{Wei Zhuo}
\affiliation{
  \institution{College of Computing and Data Science\\Nanyang Technological University}
  \country{Singapore}
}

\author{Yingbo Zhou}
\affiliation{
  \institution{  Shanghai Key Laboratory of Data Science\\Fudan University}
  \city{Shanghai}
  \country{China}
}

\author{Fan Zhang}
\affiliation{%
  \institution{ GZHU-SCHB Intelligent Transportation Joint Lab\\Guangzhou University}
  \city{Guangzhou}
  \country{China}}

\author{Siqiang Luo}
\affiliation{
  \institution{College of Computing and Data Science\\Nanyang Technological University}
  \country{Singapore}
}

\renewcommand{\shortauthors}{Li et al.}


\begin{abstract}
Spatio-temporal forecasting is a cornerstone of logistics, urban planning, and intelligent transportation systems. However, constrained by deployment costs and maintenance resources, sensor networks often lack comprehensive spatial coverage, rendering \textbf{\underline{F}}orecast \textbf{\underline{U}}nobserved \textbf{\underline{N}}ode \textbf{\underline{S}}tates (\textbf{FUNS}) a critical yet formidable challenge. Conventional models rely on historical observations and typically falter when encountering nodes without prior records. To address this, we redefine the problem as a conditional generation task on spatio-temporal graphs and propose \textbf{GenST}, a framework that introduces Large Language Models (LLMs) as a semantic bridge, leveraging a pre-trained LLM fine-tuned to extract rich semantic features from node descriptions, such as functional zones and road network structures, to compensate for missing spatio-temporal signals. Specifically, we design a two-stage generative architecture: a Spatio-Temporal VAE first compresses spatio-temporal dynamics into a latent space, followed by a Generative Transformer (GenT) that reconstructs the future states of unobserved nodes from noise, guided by multi-modal conditions including semantics, geographic coordinates, and neighborhood contexts. Experiments on six traffic and two non-traffic datasets show GenST significantly outperforms existing baselines in zero-shot prediction tasks, demonstrating the practical potential of semantic-guided generation for mitigating spatio-temporal data sparsity.
\end{abstract}




\maketitle

\section{Introduction}
\begin{figure}[t]
% \vspace{-0.15cm}
\centerline{\includegraphics[width=0.48\textwidth]{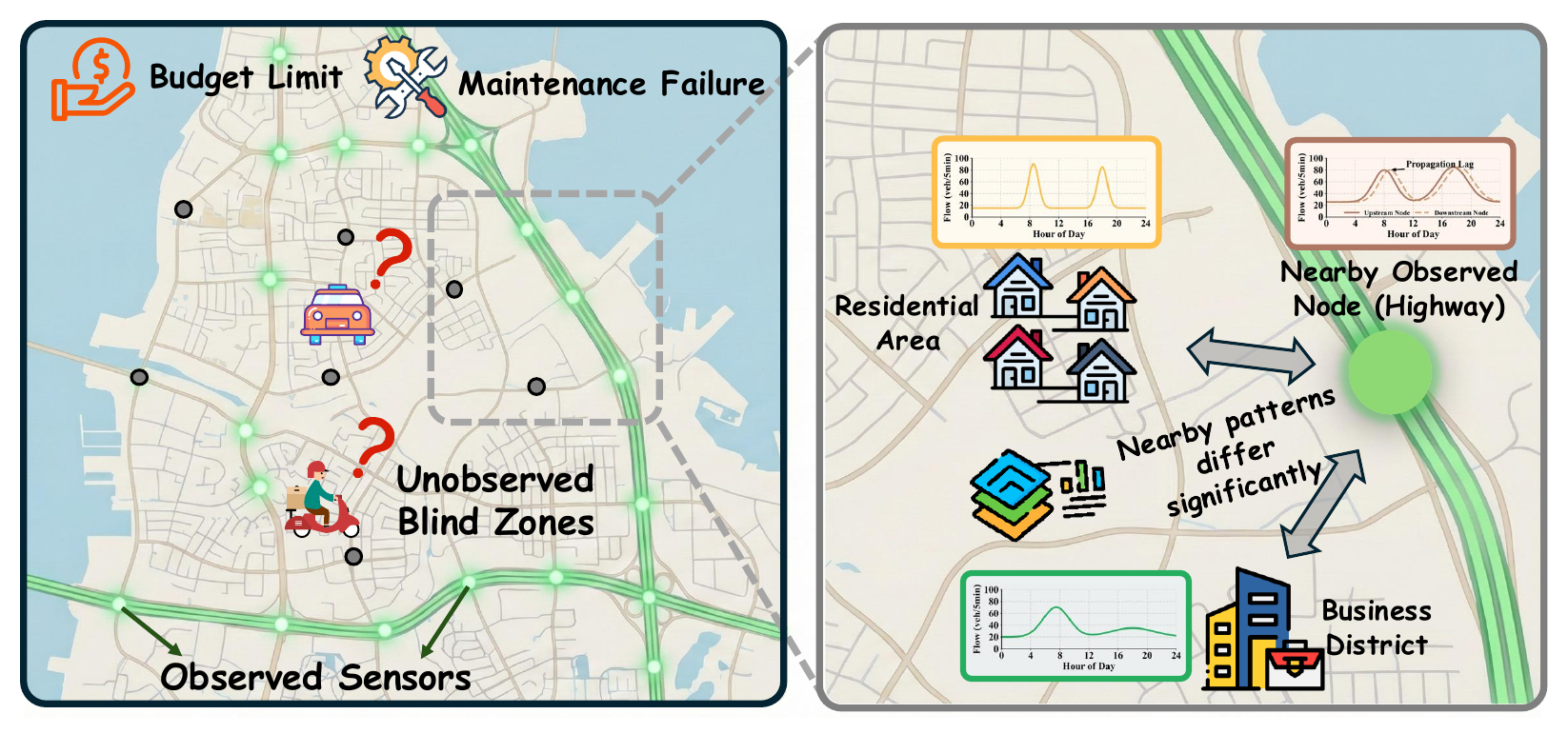}}
\vspace{-0.3cm}
\caption{Illustrative scenarios and spatial heterogeneity of unobserved nodes for FUNS.}
\vspace{-0.7cm}
\label{fig: intro}
\end{figure}

The widespread adoption of the Internet of Things (IoT) and sensor devices has established spatio-temporal data mining as a core driver of modern urban governance \cite{amato2020novel,aryandoust2022enhanced}. Accurate spatio-temporal forecasting is essential for applications such as traffic flow management \cite{li2024st,li2018diffusion}, logistics routing \cite{zhao2024just}, and energy distribution \cite{wang2024timemixer}. Existing spatio-temporal graph neural networks (STGNNs) have achieved significant success in capturing complex spatial dependencies and temporal dynamics by coupling Graph Convolutional Networks (GCNs) with sequence models like RNNs or Transformers\cite{bai2020adaptive,liu2023spatio}. However, these models are primarily designed under a transductive setting, which assumes that the set of nodes is fixed and that every node possesses complete historical observations. In essence, these models function by mapping the historical patterns of known sensors to their future states. Despite the strong performance of STGNNs, real-world deployment reveals a more challenging reality, as illustrated in \textbf{Figure~\ref{fig: intro}}: incomplete sensor coverage. High deployment costs, maintenance difficulties, and physical constraints often prevent sensor networks from covering every urban area, resulting in numerous unobserved nodes. Additionally, urban expansion frequently introduces new road segments without historical records, leading to a cold-start problem. The task of predicting future states for nodes without any initial historical observations is defined as Forecasting Unobserved Node States (FUNS). Unlike traditional data imputation, FUNS requires inductive generalization—the ability to learn universal spatio-temporal patterns from observed nodes and transfer them to unseen nodes in zero-shot.

Achieving high-precision zero-shot prediction under the FUNS setting remains difficult due to several fundamental obstacles. A primary issue is the \textbf{absence of temporal inductive bias}. Most spatio-temporal models are inherently auto-regressive, relying heavily on historical lag sequences to infer future trends \cite{guo2019attention, li2023dynamic}. For unobserved nodes where input history is absent, sequential modeling components such as GRUs or Attention mechanisms fail to activate due to the lack of signal, leading predictions to collapse to mean values or noise. While transfer learning is frequently employed to address data scarcity, it primarily focuses on parameter sharing or feature alignment across regions \cite{jin2022selective,mo2022cross}. The FUNS task, however, represents a form of structural zero-shot learning: optimized parameters alone cannot bridge the gap, as temporal mechanisms remain inert for unobserved nodes without historical signals. Consequently, establishing reasonable temporal trajectories from nothing is a significant hurdle.

Furthermore, the model must overcome \textbf{semantic blindness (e.g., road functions and surroundings) and spatial heterogeneity}, as physical proximity does not always imply pattern similarity. Existing inductive baselines, which generalize to unseen nodes, such as Inductive Kriging \cite{wu2021inductive}, aggregate neighbor information based primarily on geometric distance. Yet, urban spaces are highly heterogeneous; as shown in \textbf{Figure~\ref{fig: intro}}, whether an unobserved node is situated in a residential or a commercial district decisively influences its flow patterns. While some models attempt to capture this heterogeneity through static node embeddings or meta-learning \cite{guo2021learning, dong2024heterogeneity}, these approaches typically require learning a specific vector for each node—a process that falters when encountering entirely new nodes during inference. Even when incorporating coarse-grained priors like road types, these are often limited by hard-coded or one-hot formats that lack the capacity to describe complex, non-linear functional zones. Therefore, providing precise semantic profiles for anonymous unobserved nodes without training signal feedback is essential for resolving spatial heterogeneity.

Finally, a conflict exists between \textbf{deterministic and generative diversity}. High levels of uncertainty characterize unobserved node states. Traditional regression models attempt point estimation, which often yields overly smooth results that fail to capture the random fluctuations and peak characteristics of traffic flow \cite{marisca2024graph,wu2025spatio}. In reality, inferring unobserved states is more appropriately framed as a conditional generation task rather than a simple regression problem \cite{li2025fine}. This requires the modeling of latent probability distributions rather than merely fitting the mean values, ensuring the generated trajectories remain both diverse and physically plausible.

To address these challenges, we reformulate FUNS as a semantic-guided spatio-temporal graph generation task. Our core insight is that textual descriptions incorporating functional zones and road hierarchies can effectively replace missing numerical history as conditioning signals. LLMs, with their world knowledge, can interpret such descriptions to infer that, for instance, a commercial-district arterial road is prone to morning rush peaks, thereby providing zero-shot physical priors. Based on this, we propose the GenST framework. Rather than treating the LLM as a simple feature extractor, we introduce a Semantic Alignment Module to align the model’s semantic understanding of urban scenarios with the spatio-temporal forecasting space. Furthermore, we develop a generative architecture based on latent diffusion. A Spatio-Temporal VAE (ST-VAE) first compresses complex dynamics into a latent space, followed by a Generative Transformer (GenT) that reconstructs the trajectories of unobserved nodes from noise. This reconstruction is guided by multi-modal conditions, specifically semantic conditions and spatial neighborhood contexts. This LLM-guided generation paradigm not only resolves historical data gaps but also integrates cross-modal information, equipping the model with knowledge of urban functional structures and enabling robust and accurate zero-shot spatio-temporal forecasting in real-world scenarios.

In summary, our primary contributions are as follows:
\begin{itemize}[leftmargin=*,topsep=-3pt]
    \item We reformulate FUNS as a semantic-guided generative task on spatio-temporal graphs, providing a novel perspective for addressing cold-start challenges in graph-based forecasting.
    \item We design a semantic branch using LoRA-fine-tuned LLMs to extract open-world knowledge from road descriptions, aligning these features into condition vectors for zero-shot guidance.
    \item We propose GenST, which integrates ST-VAE and GenT to model unobserved node distributions in latent space, ensuring physically consistent state synthesis.
    \item Extensive experiments on six traffic benchmarks and two non-traffic datasets demonstrate the effectiveness and cross-domain potential of GenST under zero-shot unobserved-node forecasting, with robust performance under sparse observations.
\end{itemize}

\section{Preliminaries}

\begin{figure}[t]
\vspace{-0.15cm}
% \hspace{0.9cm}
\centerline{\includegraphics[width=0.5\textwidth]{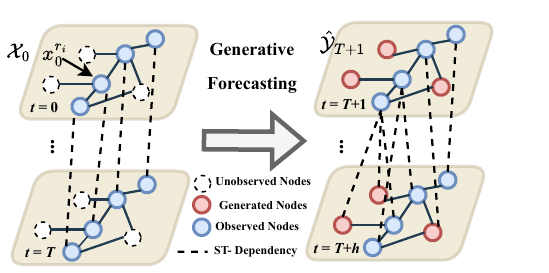}}
\vspace{-0.3cm}
\caption{Reformulating FUNS as unobserved node generation based on observed graphs.}
\vspace{-0.6cm}
\label{fig: problem}
\end{figure}
In this section, we define the notations and provide a formal description of FUNS, complemented by the illustration in \textbf{Figure~\ref{fig: problem}}.

\noindent\textbf{\textit{Definition 1 (Spatio-temporal Graph)}}: We represent an urban road network as a spatio-temporal graph $G = (V, E, A)$, where $V = \{r_1, r_2, \dots, r_N\}$ is a set of $N$ road segment nodes, and the edge set $E$ represents physical connections or proximity between segments. The adjacency matrix $A \in \mathbb{R}^{N \times N}$ characterizes the spatial topology, where $A_{ij} > 0$ indicates a spatial correlation between road segments $r_i$ and $r_j$. Under the FUNS setting, the node set $V$ is partitioned into two mutually exclusive subsets $V^+$ and $V^-$, such that $V = V^+ \cup V^-$ and $N = N^+ + N^-$, where $N^+$ and $N^-$ represent the number of observed and unobserved nodes, respectively.

\noindent\textbf{\textit{Definition 2 (Observed Nodes)}}: The set of observed nodes is denoted as $V^+ = \{o_1, o_2, \dots, o_{N^+} \} \subset V$, representing road segments equipped with sensors that provide continuous historical observations. These measurements form the basis for the model to learn universal spatio-temporal evolution patterns.

\noindent\textbf{\textit{Definition 3 (Unobserved Nodes)}}: The set of unobserved nodes is denoted as $V^- = \{u_1, u_2, \dots, u_{N^-} \} = V \setminus V^+$, representing road segments without sensors. Historical data for these nodes is entirely missing during both the training and testing phases.

\begin{figure*}[t]
\vspace{-0.4cm}
	\centering
		\includegraphics[width=0.7\linewidth]{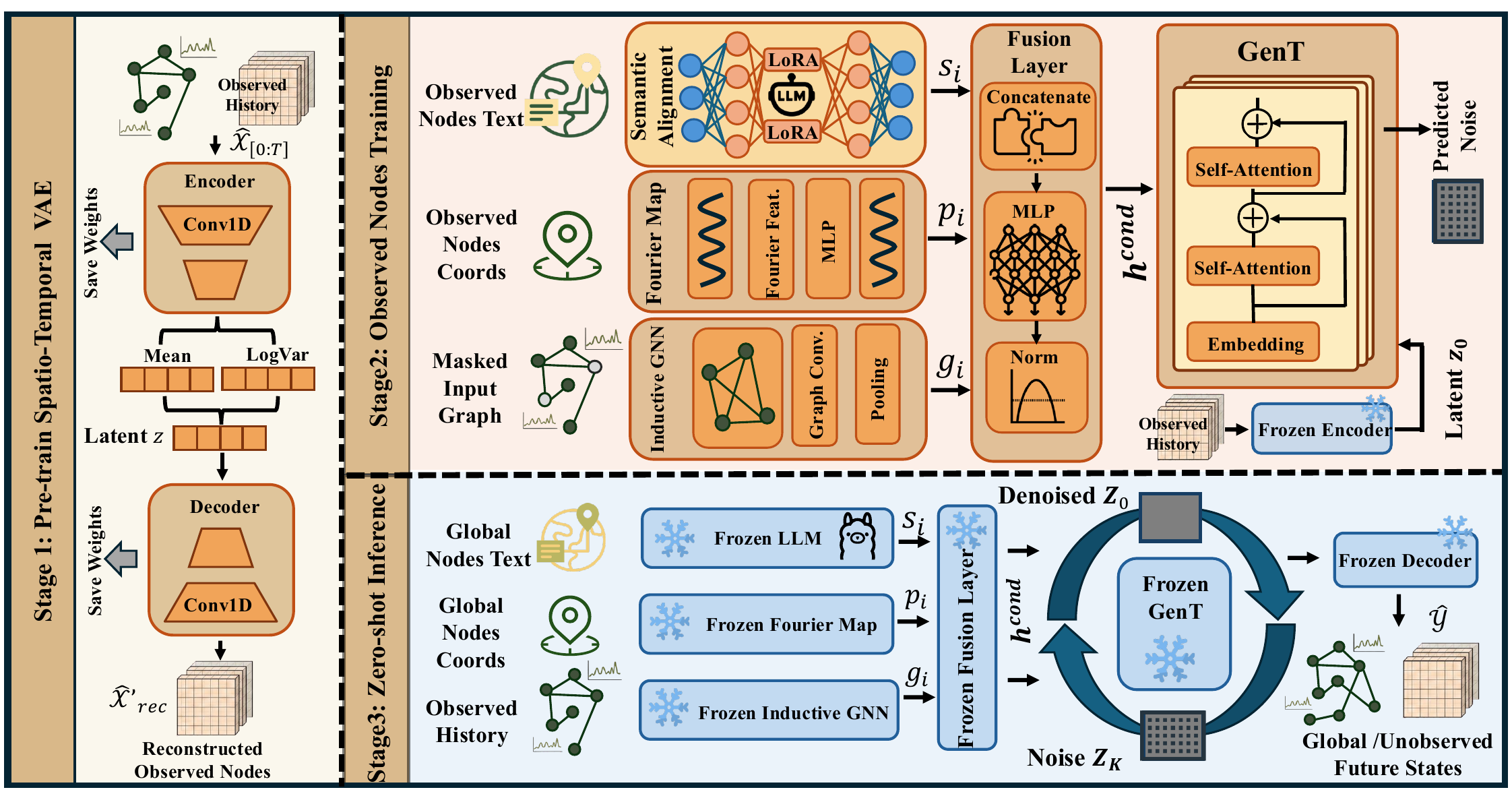}
  \vspace{-0.3cm}
		\caption{The GenST architecture. The architecture features a unified generative pipeline centered on Stage 2, where latent foundation is unified with multi-modal context alignment to drive the primary engine for direct Stage 3 zero-shot inference.}
		\label{fig: GenST}
  \vspace{-0.4cm}
\end{figure*}

To represent traffic states, we let $x^{r_i}_t \in \mathbb{R}^D$ represent the average traffic state (e.g., flow or speed) of road segment $r_i$ at time $t$, where $D$ is the feature dimension. The state of the entire network at time $t$ is expressed as $\mathcal{X}_t = [x_t^{r_1}, x_t^{r_2}, \dots, x_t^{r_N}] \in \mathbb{R}^{N \times D}$. The spatio-temporal tensor for the full graph across $T$ time steps is denoted as $\mathcal{X}_{[0:T]} \in \mathbb{R}^{T \times N \times D}$. During both the training and inference phases, the model only has access to the historical sequences of observed nodes as input, denoted as $\hat{\mathcal{X}}_{[0:T]} \in \mathbb{R}^{T \times N^+ \times D}$.

\noindent\textbf{\textit{Definition 4 (Textual Semantics)}}: Textual semantics refers to the text description $d_i$ associated with each node $r_i$ (e.g., describing its functional zone or road category). These qualitative descriptions are transformed into high-dimensional semantic feature vectors $\mathbf{s}_i \in \mathbb{R}^{d_{emb}}$, forming the global semantic feature matrix $S \in \mathbb{R}^{N \times d_{emb}}$.

\noindent\textbf{\textit{Definition 5 (Geographic Coordinates)}}: Geographic coordinates refer to the longitude and latitude $(lon_i, lat_i)$ corresponding to each node $r_i$, denoted as $c_i$. The resulting matrix $C \in \mathbb{R}^{N \times 2}$ provides the model with absolute position priors in continuous space.

\noindent\textbf{\textit{Problem Definition (FUNS)}}: The objective of Forecasting Unobserved Node States is to learn a mapping function $f$. Given the historical sequences of observed nodes $\hat{\mathcal{X}}_{[0:T]}$, the global graph structure $G$, geographic coordinates $C$, and textual semantics $S$, the task typically involves the following two formal objectives:

\noindent(1) \textbf{Global Network Forecasting:} Predicting the future states of the entire road network, including both observed and unobserved nodes, for the next $h$ time steps:
\begin{equation}
\hat{\mathcal{Y}} = f(\hat{\mathcal{X}}_{[0:T]}, G, C, S) \in \mathbb{R}^{h \times N \times D}.
\end{equation}
\noindent(2) \textbf{Unobserved Node Forecasting:} Specifically predicting the states of the unobserved node set $V^-$, which lacks any historical records, for the next $h$ time steps:
\begin{equation}
\hat{\mathcal{Y}}^- = f(\hat{\mathcal{X}}_{[0:T]}, G, C, S) \in \mathbb{R}^{h \times N^- \times D}.
\end{equation}
FUNS differs from traditional data imputation \cite{ke2025geomae,yang2025graph}, which primarily addresses random, short-term data missingness caused by sensor failures. Instead, FUNS is categorized as inductive zero-shot learning, requiring the model to synthesize future spatio-temporal trajectories for target nodes entirely lacking historical records by leveraging spatial correlations and semantic priors.

\section{Methodology}

To overcome missing temporal biases and spatial heterogeneity in FUNS, we introduce GenST, reformulating forecasting as a multi-modal conditional graph node generation task, as illustrated in \textbf{Figure~\ref{fig: GenST}}. The framework orchestrates a cohesive pipeline: a ST-VAE first distills intricate traffic dynamics into a compact latent manifold, providing a robust foundation for our core engine—the GenT. During training, GenT masters denoising and reconstruction patterns guided by multi-modal priors, including LoRA-fine-tuned semantics, coordinates, and neighborhood topology. Ultimately, GenST leverages these learned distributions to synthesize future states for the entire network zero-shot. The following sections detail this architecture, prioritizing the pivotal generative stage.

\subsection{Multi-modal Context Alignment}
Since unobserved nodes $V^-$ lack historical sequences $\hat{\mathcal{X}}_{[0:T]}$, autoregressive mechanisms fail to capture temporal dependencies. We bridge this critical gap via multi-modal context alignment, which distills open-world semantics and structural priors from text descriptions, coordinates, and topology into a unified space to generate guidance conditions $\mathbf{h}^{cond}$ for zero-shot inference.

\subsubsection{Semantic Alignment with LLMs}
Road segments with identical physical attributes can exhibit diverging traffic patterns depending on their functional zones. To capture this semantic logic, we harness an LLM as an interpreter. For each node $r_i$, we obtain text description $d_i$ (\textbf{Definition 4}) and employ LoRA fine-tuning—freezing the pre-trained weights $\Theta_{LLM}$ and injecting low-rank trainable matrices $\mathbf{B} \in \mathbb{R}^{d \times r}$ and $\mathbf{A} \in \mathbb{R}^{r \times d}$ into each attention layer—to adapt the LLM's knowledge to traffic forecasting. Concretely, for each adapted attention layer, the original weight matrix $\mathbf{W}_0$ is augmented as $\mathbf{W} = \mathbf{W}_0 + \frac{\alpha}{r}\mathbf{B}\mathbf{A}$, where $\alpha$ is a scaling factor and $r$ is the LoRA rank. The semantic embedding $\mathbf{s}_i$ is then computed as:
\begin{equation}
\mathbf{s}_i = \text{LLM}(d_i; \Theta_{LLM}, \Theta_{LoRA}) = \text{Proj}(\mathbf{o}_{last}),
\end{equation}
where $\mathbf{o}_{last}$ is the hidden state of the final LLM layer produced under the LoRA-adapted parameters $\{\mathbf{W}_0 + \frac{\alpha}{r}\mathbf{B}\mathbf{A}\}$, and $\text{Proj}(\cdot)$ is a linear layer that maps the LLM's hidden dimension to the model feature space $d_{model}$. Through this process, the unstructured text $d_i$ is transformed into a continuous vector $\mathbf{s}_i$ containing deep functional semantics.
\subsubsection{Spatial and Structural Encoding}
In addition to semantic information, absolute geographic locations and local topological structures are vital for inductive inference on $V^-$.

\noindent\textbf{Geographic Coordinate Encoding}:
Inductive inference on $V^-$ requires absolute locations to resolve position biases inherent to urban space. Following \textbf{Definition 5}, we extract coordinates $c_i \in C$ and apply Fourier Feature Mapping $\gamma(\cdot)$, yielding
\begin{equation}
\mathbf{p}_i = \gamma(c_i) = [\cos(2\pi \mathbf{W}_{freq} c_i), \sin(2\pi \mathbf{W}_{freq} c_i)], 
\end{equation} 
where $\mathbf{W}_{freq}$ is a learnable Gaussian random frequency matrix. The resulting absolute position embedding $\mathbf{p}_i$ enables the model to capture the geographic distribution of unobserved nodes $u_j \in V^-$.

\noindent\textbf{Inductive Structural Encoding}:
To capture local topology and dynamic contexts, we architect an Inductive GNN. Unlike ID-based methods that falter zero-shot, this module performs message passing using the adjacency matrix $A$, static attributes, and neighbor historical encodings. Specifically, for each observed neighbor $o_j \in V^+$, we obtain its historical encoding $\mathbf{z}_j^{hist} = E_\phi(\hat{\mathcal{X}}_{[0:T]}^{o_j}) \in \mathbb{R}^{d_{latent}}$ by feeding its historical sequence into the frozen ST-VAE encoder. The structural feature is then computed via graph convolution:
\begin{equation}
\mathbf{g}_i = \sigma \left( \sum_{j \in \mathcal{N}(i)} \frac{1}{\sqrt{\tilde{d}_i \tilde{d}_j}} \mathbf{W}_{gnn} \cdot \text{Concat}(\mathbf{f}_j^{s}, \mathbf{z}_j^{hist}) \right),
\end{equation}
where $\mathbf{f}_j^{s} \in \mathbb{R}^{d_s}$ denotes the static attributes of neighbor $j$, $\tilde{d}_i = 1 + \sum_j A_{ij}$ is the augmented degree, and $\mathbf{W}_{gnn}$ is a learnable weight matrix. For unobserved neighbors $u_k \in V^-$ whose historical are unavailable, a learnable mask token $\mathbf{m} \in \mathbb{R}^{d_{latent}}$ substitutes $\mathbf{z}_k^{hist}$, allowing node $i$ to generate context-aware features $\mathbf{g}_i$ by aggregating neighborhood dynamics even under partial observability.

\subsubsection{Context Fusion}
To provide a unified guidance signal for the generative model, we concatenate the three heterogeneous features—semantic embeddings $\mathbf{s}_i$, absolute position embeddings $\mathbf{p}_i$, and structural features $\mathbf{g}_i$—along the feature dimension and fuse them adaptively using a Fusion MLP:
\begin{equation}
\mathbf{h}^{cond}_i = \text{LayerNorm}\left( \text{MLP}_{fusion}([\mathbf{s}_i \parallel \mathbf{p}_i \parallel \mathbf{g}_i]) \right).
\end{equation}
The resulting $\mathbf{h}^{cond}_i \in \mathbb{R}^{d_{model}}$ serves as
Key and Value for Cross-Attention in GenT, guiding latent denoising.

\subsection{Spatio-Temporal VAE}
Direct diffusion in the high-dimensional, sparse raw space $\hat{\mathcal{X}}_{[0:T]}$ is computationally prohibitive and struggles to capture complex dependencies \cite{mei2025deep,yun2025posterior}. We introduce the ST-VAE to learn a continuous, physically meaningful latent manifold. This compact space encodes intrinsic traffic evolution patterns, significantly reducing the complexity of downstream generative tasks.
\subsubsection{Spatio-Temporal Encoder}
To distill temporal dependencies, we design a lightweight Conv1D-based encoder to infer the posterior $q_\phi(z|\hat{\mathcal{X}}_{[0:T]})$. After spatial flattening, multi-layer Conv1Ds extract features to parameterize a Gaussian distribution:
\begin{equation}
\mu = \text{Conv1D}{\mu}(\hat{\mathcal{X}}_{[0:T]}), \quad \log \sigma^2 = \text{Conv1D}{\sigma}(\hat{\mathcal{X}}_{[0:T]}),
\end{equation}
using the reparameterization trick, the latent variable $z \in \mathbb{R}^{N^+ \times d_{latent}}$ is sampled as follows:
\begin{equation}
z = \mu + \sigma \odot \epsilon, \quad \epsilon \sim \mathcal{N}(0, \mathbf{I}),
\end{equation}
where $\epsilon$ denotes standard Gaussian noise and $d_{latent}$ represents the dimension of the latent space. This process compresses complex spatio-temporal dynamics into a low-dimensional latent vector $z$, which serves as the generation target for GenT.
\subsubsection{Spatio-Temporal Decoder}
The decoder $p_\theta(\hat{\mathcal{X}}_{[0:T]}|z)$ is designed to verify whether the latent variable $z$ retains sufficient original information to reconstruct the observed history. Its structure is symmetric to the encoder, mapping the latent vector $z$ back to the original spatio-temporal dimensions via transposed 1D convolutions:
\begin{equation}
\hat{\mathcal{X}}'_{rec} = \text{Decoder}(z) \in \mathbb{R}^{T \times N^+ \times D}.
\end{equation}
Through this reconstruction mechanism, we force the latent space to preserve key physical characteristics of traffic flow, such as periodicity and seasonality, rather than simply fitting noise.
\subsubsection{Variational Objective}
The training of the Spatio-Temporal VAE follows the principle of maximizing the evidence lower bound. The total loss function $\mathcal{L}_{VAE}$ consists of a reconstruction loss and a regularization term:
\begin{equation}
\begin{split}
    \mathcal{L}_{VAE} &= \mathbb{E}_{q_\phi(z|\hat{\mathcal{X}}_{[0:T]})} \left[ \|\hat{\mathcal{X}}_{[0:T]} - \hat{\mathcal{X}}'_{rec}\|^2_2 \right] \\
    &\quad + \lambda \cdot D_{KL} \left( q_\phi(z|\hat{\mathcal{X}}_{[0:T]}) \parallel \mathcal{N}(0, \mathbf{I}) \right).
    \label{eq:vae_loss}
\end{split}
\end{equation}
The reconstruction term ensures fidelity, while the KL divergence regularizes the posterior toward a standard normal distribution, with $\lambda$ balancing this trade-off.

\subsection{Generative Transformer}
By leveraging the compact and continuous latent manifold constructed via ST-VAE and rich guidance from context alignment, we designed the core generative engine: GenT. As a semantic-guided generator, GenT reformulates forecasting as a conditional probabilistic process on the manifold. By employing Spatial-Semantic Cross-Attention, it dynamically transforms static semantics $\mathbf{s}_i$, coordinates $\mathbf{p}_i$, and topology into temporal evolution dynamics, enabling precise zero-shot synthesis for unobserved nodes.

\subsubsection{Latent Diffusion Process}
The generation process is established on the latent manifold $\mathbf{z}_0 = E_\phi(\mathcal{Y})$ extracted by the VAE encoder $E_\phi$ from the future ground truth, circumventing high-frequency noise and focusing on high-level structural semantics. Note that although $E_\phi$ is pre-trained on historical sequences, reusing it to encode future windows is well-justified, and $E_\phi$ serves solely as a domain-agnostic temporal compressor providing the latent space—the actual forecasting is entirely delegated to GenT.

\noindent\textbf{Manifold Diffusion}: We model traffic state degradation as a Markov chain by progressively injecting Gaussian noise into the latent representation $\mathbf{z}_0$. For steps $\tau \in \{1, \ldots, K\}$, the transition probability is:
\begin{equation}
q(\mathbf{z}_\tau | \mathbf{z}_{\tau-1}) = \mathcal{N}(\mathbf{z}_\tau; \sqrt{1 - \beta_\tau} \mathbf{z}_{\tau-1}, \beta_\tau \mathbf{I}).
\end{equation}

As $\tau$ increases, structured features dissolve until $\mathbf{z}_K$ approximates isotropic Gaussian noise $\mathcal{N}(0, \mathbf{I})$.

\noindent\textbf{Conditional Trajectory Reconstruction}:
Generation learns the reverse transformation via a network $\epsilon_\theta$ that approximates the posterior $p_\theta(\mathbf{z}_{\tau-1} | \mathbf{z}_\tau, \mathbf{h}^{cond})$ under multi-modal guidance:
\begin{equation}
p_\theta(\mathbf{z}_{\tau-1} | \mathbf{z}_\tau, \mathbf{h}^{cond}) = \mathcal{N}(\mathbf{z}_{\tau-1}; \boldsymbol{\mu}_\theta(\mathbf{z}_\tau, \tau, \mathbf{h}^{cond}), \sigma_\tau^2 \mathbf{I}).
\end{equation}

The core challenge is designing $\boldsymbol{\mu}_\theta$ to transform static conditions $\mathbf{h}^{cond}$ into a dynamic force field that guides the manifold walk, which is the focus of the GenT architecture.
\subsubsection{Structure-Aware Generative Backbone}
Traditional diffusion models often employ convolution-based U-Net architectures \cite{ma2025efficient,cao2024survey}, which struggle with non-Euclidean graph structures and long-range spatio-temporal dependencies. Therefore, we designed the Structure-Aware Generative Transformer as the denoising backbone. This architecture discards the limitations of local inductive bias and achieves a unified capture of global spatio-temporal correlations and multi-modal knowledge injection by alternately stacking Self-Attention and Cross-Attention layers.

\noindent\textbf{Self-Attention}: At any step $\tau$, the input $\mathbf{z}_\tau$ passes through a Self-Attention layer to establish consistency across generated trajectories, ensuring coherent spatio-temporal flow propagation:
\begin{equation}
\mathbf{H}_{self} = \text{Attention}(\mathbf{Q}_z, \mathbf{K}_z, \mathbf{V}_z) = \text{softmax}\left(\frac{\mathbf{Q}_z\mathbf{K}_z^T}{\sqrt{d_{head}}}\right)\mathbf{V}_z,
\end{equation}
where the Query, Key, and Value are linear projections of $\mathbf{z}_\tau$. This mechanism allows the model to maintain a global spatio-temporal dependency map at every denoising step.

\noindent\textbf{Cross-Attention}: Serving as a knowledge injector, this module connects static semantics to dynamic states. Queries derived from $\mathbf{H}_{self}$ interact with Keys and Values projected from $\mathbf{h}^{cond}$ :
\begin{equation} \mathbf{Q} = \mathbf{H}_{self} \mathbf{W}_Q, \quad \mathbf{K} = \mathbf{h}^{cond} \mathbf{W}_K, \quad \mathbf{V} = \mathbf{h}^{cond} \mathbf{W}_V, \end{equation} 
\begin{equation} \mathbf{H}_{cross} = \text{Attention}(\mathbf{Q}, \mathbf{K}, \mathbf{V}).
\end{equation}

By computing similarity, the model adaptively retrieves semantic prototypes and spatial priors from $\mathbf{h}^{cond}$ based on the current generation state. This mechanism effectively translates static text and coordinates into dynamic evolution rules, filling the information vacuum left by the absence of numerical history.
\subsubsection{Semantics-Consistent Denoising Objective}
GenT optimizes for a semantics-consistent conditional generative distribution rather than mere signal recovery. We employ a weighted mean squared error as the loss function, forcing the model to predict the noise $\epsilon$ added to the latent variable $\mathbf{z}_0$:
\begin{equation} \mathcal{L}_{diff} = \mathbb{E}{\mathbf{z}_0, \tau, \epsilon, \mathbf{h}^{cond}} \left[ || \epsilon - \epsilon_\theta(\mathbf{z}_\tau, \tau, \mathbf{h}^{cond}) ||^2_2 \right]. 
\end{equation}

By minimizing $\mathcal{L}_{diff}$, GenT learns to restore the latent manifold $\mathbf{z}_0$ representing real traffic states from pure Gaussian noise under the strong guidance of multi-modal conditions $\mathbf{h}^{cond}$.

\subsection{Zero-Shot Inference}
During the inference phase, the objective is to directly synthesize future states $\hat{\mathcal{Y}}$ for the entire network, including unobserved nodes. Although training is conducted solely on the observed subgraph ($N^+$), GenST’s fully inductive design allows it to generalize to global graph structures of any scale ($N$) without any fine-tuning. This is achieved by completely discarding node-specific ID embeddings in favor of attention mechanisms that are insensitive to input length, thereby learning universal spatio-temporal evolution rules.

The inference process follows a generative sampling paradigm. We first initialize pure Gaussian noise $\mathbf{z}_K \in \mathbb{R}^{N \times d_{latent}}$ across the entire network. Subsequently, using the global multi-modal conditions $\mathbf{h}^{cond}$ which incorporate semantics $\mathbf{s}_i$, coordinates $\mathbf{p}_i$, and structural features $\mathbf{g}_i$ for unobserved nodes, we perform a reverse denoising cycle. For each step $\tau = K, K-1, \dots, 1$, the model updates the network state using learned physical rules:
\begin{equation}
\mathbf{z}_{\tau-1} = \frac{1}{\sqrt{\alpha_\tau}} \left( \mathbf{z}_\tau - \frac{1-\alpha_\tau}{\sqrt{1-\bar{\alpha}_\tau}} \epsilon_\theta(\mathbf{z}_\tau, \tau, \mathbf{h}^{cond}) \right) + \sigma_\tau \boldsymbol{\xi},
\end{equation}
where $\boldsymbol{\xi} \sim \mathcal{N}(0, \mathbf{I})$, $\alpha_\tau = 1 - \beta_\tau$ and $\bar{\alpha}_\tau = \prod_{s=1}^\tau \alpha_s$. This process essentially sculpts disordered global noise into physically meaningful latent trajectories $\mathbf{z}_0$ under the guidance of the global condition field $\mathbf{h}^{cond}$.
Finally, by mapping back to the original space via the frozen VAE decoder, we obtain the complete future state prediction for the road network:
\begin{equation}
\hat{\mathcal{Y}} = D_\psi(\mathbf{z}_0) \in \mathbb{R}^{h \times N \times D}.
\end{equation}

This whole-graph generation approach enables zero-shot inference for unobserved regions $V^-$ and ensures the physical consistency of predictions across the entire network topology.

\section{Experiments}
\begin{table*}[t]
\vspace{-0.5cm}
  \centering
  \caption{Main performance comparison. Best Global/Unobserved results are highlighted in bold with dark blue and dark orange colors, respectively; second-best results are indicated with light blue and light orange backgrounds.}
  \vspace{-0.4cm}
  \resizebox{0.95\linewidth}{!}{
  \setlength{\tabcolsep}{3.5pt}
    \begin{tabular}{cccccccccccccccccccc}
    \toprule
    \multirow{2}[4]{*}{\textbf{Models}} & \textbf{Datasets} & \multicolumn{3}{c}{\textbf{METR-LA}} & \multicolumn{3}{c}{\textbf{PeMS-Bay}} & \multicolumn{3}{c}{\textbf{PeMS03}} & \multicolumn{3}{c}{\textbf{PeMS04}} & \multicolumn{3}{c}{\textbf{PeMS07}} & \multicolumn{3}{c}{\textbf{PeMS08}} \\
\cmidrule{2-20}          & \textbf{Tasks} & \textbf{MAE} & \textbf{RMSE} & \textbf{MAPE} & \textbf{MAE} & \textbf{RMSE} & \textbf{MAPE} & \textbf{MAE} & \textbf{RMSE} & \textbf{MAPE} & \textbf{MAE} & \textbf{RMSE} & \textbf{MAPE} & \textbf{MAE} & \textbf{RMSE} & \textbf{MAPE} & \textbf{MAE} & \textbf{RMSE} & \textbf{MAPE} \\
    \midrule
    \multirow{2}[1]{*}{\textbf{K-NN}} & Global & 16.84  & 26.76  & 30.38\% & 13.79  & 19.05  & 27.50\% & 81.28  & 124.99  & 91.34\% & 107.81  & 141.79  & 62.36\% & 107.62  & 176.30  & 79.96\% & 101.17  & 135.04  & 148.28\% \\
          & Unobserved & 20.90  & 33.01  & 35.72\% & 14.83  & 20.62  & 34.78\% & 147.77  & 230.74  & 275.44\% & 180.46  & 247.05  & 136.68\% & 243.90  & 323.19  & 268.02\% & 194.22  & 259.33  & 316.14\% \\
    \multirow{2}[0]{*}{\textbf{Kriging}} & Global & 15.50  & 21.09  & 28.21\% & 13.21  & 22.43  & 28.42\% & 82.89  & 124.68  & 92.26\% & 99.81  & 128.27  & 82.80\% & 132.57  & 159.28  & 141.37\% & 107.12  & 141.78  & 170.59\% \\
          & Unobserved & 18.82  & 31.07  & 32.41\% & 14.77  & 24.03  & 29.48\% & 116.74  & 141.40  & 210.65\% & 122.15  & 157.50  & 144.07\% & 165.09  & 206.04  & 170.74\% & 145.50  & 181.34  & 297.98\% \\
    \multirow{2}[1]{*}{\textbf{HA}} & Global & 20.97  & 24.87  & 32.56\% & 12.44  & 22.43  & 27.26\% & 92.78  & 116.64  & 214.90\% & 111.10  & 138.07  & 165.64\% & 133.26  & 162.86  & 150.08\% & 98.42  & 122.49  & 123.92\% \\
          & Unobserved & 25.45  & 36.64  & 37.41\% & 13.91  & 24.03  & 28.28\% & 108.28  & 132.28  & 410.08\% & 135.97  & 169.53  & 174.18\% & 165.95  & 198.23  & 289.96\% & 133.68  & 156.67  & 278.48\% \\
\cmidrule{2-20}    \multirow{2}[2]{*}{\textbf{Mean}} & Global & 17.77  & 24.24  & 30.38\% & 13.15  & 21.30  & 27.73\% & 85.65  & 122.10  & 132.83\% & 106.24  & 136.04  & 103.60\% & 124.48  & 166.15  & 123.80\% & 102.24  & 133.10  & 147.60\% \\
          & Unobserved & 21.72  & 33.57  & 35.18\% & 14.50  & 22.89  & 30.85\% & 124.26  & 168.14  & 298.72\% & 146.19  & 191.36  & 151.64\% & 191.65  & 242.49  & 242.91\% & 157.80  & 199.11  & 297.53\% \\
    \midrule
    \midrule
    \multirow{2}[1]{*}{\textbf{STGCN}} & Global & 13.96  & 19.65  & 25.12\% & 7.42  & 11.69  & 18.81\% & 32.73  & 56.01  & 49.12\% & 37.57  & 62.65  & 27.43\% & 42.36  & 66.12  & 28.62\% & 34.14  & 56.67  & 39.47\% \\
          & Unobserved & 15.31  & 22.36  & 28.81\% & 7.81  & 12.07  & 20.48\% & 78.31  & 107.81  & 153.65\% & 84.20  & 116.97  & 58.69\% & 83.47  & 114.37  & 74.21\% & 83.92  & 109.69  & 135.81\% \\
    \multirow{2}[0]{*}{\textbf{GWNet}} & Global & 13.13  & 18.49  & 23.63\% & 7.46  & 11.76  & 18.92\% & 30.33  & 51.90  & 45.52\% & 35.27  & 58.81  & 25.75\% & 39.19  & 61.17  & 26.48\% & 31.12  & 51.66  & 35.98\% \\
          & Unobserved & 14.38  & 20.50  & 26.75\% & 8.03  & 12.40  & 21.17\% & 74.49  & 101.07  & 143.52\% & 81.27  & 111.10  & 56.98\% & 79.16  & 107.50  & 66.96\% & 79.38  & 101.18  & 126.25\% \\
    \multirow{2}[0]{*}{\textbf{AGCRN}} & Global & 14.16  & 21.88  & 26.21\% & 7.17  & 11.41  & 18.31\% & 31.20  & 54.73  & 49.23\% & 36.13  & 62.02  & 25.77\% & 42.11  & 66.53  & 31.51\% & 32.21  & 55.86  & 36.83\% \\
          & Unobserved & 15.48  & 23.62  & 29.29\% & 7.88  & 12.29  & 21.04\% & 78.61  & 107.81  & 156.45\% & 85.54  & 118.53  & 58.92\% & 87.13  & 118.76  & 77.66\% & 85.15  & 110.68  & 131.74\% \\
    \multirow{2}[0]{*}{\textbf{STGODE}} & Global & 13.26  & 18.67  & 23.87\% & 7.15  & 11.27  & 18.13\% & 32.01  & 54.77  & 48.04\% & 35.65  & 59.44  & 26.03\% & 38.84  & 60.63  & 26.24\% & 30.08  & 49.92  & 34.77\% \\
          & Unobserved & 14.52  & 20.70  & 27.03\% & 7.69  & 11.89  & 20.29\% & 78.62  & 106.66  & 151.47\% & 82.15  & 112.29  & 57.60\% & 78.45  & 106.55  & 66.36\% & 76.73  & 97.77  & 122.00\% \\
    \multirow{2}[0]{*}{\textbf{GTS}} & Global & 14.44  & 20.32  & 25.98\% & 7.46  & 11.76  & 18.92\% & 30.80  & 52.71  & 46.23\% & 40.51  & 67.55  & 29.58\% & 42.80  & 66.80  & 28.92\% & 31.56  & 52.39  & 36.49\% \\
          & Unobserved & 15.81  & 22.53  & 29.41\% & 8.03  & 12.40  & 21.17\% & 75.65  & 102.64  & 145.76\% & 93.35  & 127.61  & 65.46\% & 86.45  & 117.39  & 73.13\% & 80.50  & 102.61  & 128.04\% \\
    \multirow{2}[0]{*}{\textbf{STID}} & Global & 13.57  & 19.10  & 24.42\% & 7.51  & 11.83  & 19.04\% & 32.13  & 54.98  & 48.22\% & 34.65  & 57.78  & 25.30\% & 37.25  & 58.14  & 25.17\% & 30.10  & 49.96  & 34.80\% \\
          & Unobserved & 14.86  & 21.18  & 27.65\% & 8.08  & 12.48  & 21.30\% & 78.91  & 107.06  & 152.04\% & 79.85  & 109.15  & 55.99\% & 75.24  & 102.18  & 63.65\% & 76.78  & 97.85  & 122.11\% \\
    \multirow{2}[1]{*}{\textbf{MegaCRN}} & Global & 12.92  & 18.18  & 23.24\% & 7.33  & 11.55  & 18.58\% & 31.20  & 53.39  & 46.82\% & 35.84  & 59.76  & 26.16\% & 37.51  & 58.56  & 25.35\% & 30.35  & 50.38  & 35.09\% \\
          & Unobserved & 14.15  & 20.16  & 26.31\% & 7.89  & 12.18  & 20.79\% & 76.63  & 103.97  & 147.62\% & 82.59  & 112.89  & 57.89\% & 75.76  & 102.91  & 64.10\% & 77.42  & 98.67  & 123.13\% \\
\cmidrule{2-20}    \multirow{2}[2]{*}{\textbf{Mean}} & Global & 13.63  & 19.47  & 24.64\% & 7.36  & 11.61  & 18.67\% & 31.49  & 54.07  & 47.60\% & 36.52  & 61.14  & 26.57\% & 40.01  & 62.56  & 27.47\% & 31.37  & 52.41  & 36.20\% \\
          & Unobserved & 14.93  & 21.58  & 27.89\% & 7.92  & 12.24  & 20.89\% & 77.32  & 105.29  & 150.07\% & 84.14  & 115.51  & 58.79\% & 80.81  & 109.95  & 69.44\% & 79.98  & 102.64  & 127.01\% \\
    \midrule
    \midrule
    \multirow{2}[1]{*}{\textbf{STEP}} & Global & 12.10  & 17.03  & 21.76\% & \cellcolor[rgb]{ .855,  .914,  .973}5.72  & \cellcolor[rgb]{ .855,  .914,  .973}9.01  & \cellcolor[rgb]{ .855,  .914,  .973}14.50\% & 28.50  & 48.78  & 42.79\% & 31.38  & 52.34  & 22.91\% & \cellcolor[rgb]{ .855,  .914,  .973}35.15  & \cellcolor[rgb]{ .855,  .914,  .973}54.86  & 23.75\% & 28.94  & 48.03  & 33.45\% \\
          & Unobserved & 13.25  & 18.88  & 24.64\% & \cellcolor[rgb]{ .984,  .886,  .835}6.15  & \cellcolor[rgb]{ .984,  .886,  .835}9.50  & \cellcolor[rgb]{ .984,  .886,  .835}16.23\% & 70.00  & 95.00  & 134.90\% & 72.32  & 98.87  & 50.71\% & 70.98  & 96.42  & 60.06\% & 73.81  & 94.07  & 117.39\% \\
    \multirow{2}[1]{*}{\textbf{ST-LLM+}} & Global & \cellcolor[rgb]{ .855,  .914,  .973}11.56  & \cellcolor[rgb]{ .855,  .914,  .973}16.81  & \cellcolor[rgb]{ .855,  .914,  .973}21.00\% & 6.17  & 9.77  & 15.70\% & \cellcolor[rgb]{ .855,  .914,  .973}26.41  & \cellcolor[rgb]{ .855,  .914,  .973}45.56  & \cellcolor[rgb]{ .855,  .914,  .973}40.30\% & \cellcolor[rgb]{ .855,  .914,  .973}30.53  & \cellcolor[rgb]{ .855,  .914,  .973}51.40  & \cellcolor[rgb]{ .855,  .914,  .973}22.11\% & \cellcolor[rgb]{ .651,  .788,  .925}\textbf{34.64 } & \cellcolor[rgb]{ .651,  .788,  .925}\textbf{54.30 } & 24.26\% & \cellcolor[rgb]{ .855,  .914,  .973}27.31  & \cellcolor[rgb]{ .855,  .914,  .973}45.99  & \cellcolor[rgb]{ .855,  .914,  .973}31.46\% \\
          & Unobserved & \cellcolor[rgb]{ .984,  .886,  .835}12.65  & \cellcolor[rgb]{ .984,  .886,  .835}18.64  & \cellcolor[rgb]{ .984,  .886,  .835}23.77\% & 6.64  & 10.30  & 17.56\% & \cellcolor[rgb]{ .984,  .886,  .835}64.85  & \cellcolor[rgb]{ .984,  .886,  .835}88.72  & \cellcolor[rgb]{ .984,  .886,  .835}127.07\% & \cellcolor[rgb]{ .984,  .886,  .835}70.34  & \cellcolor[rgb]{ .984,  .886,  .835}97.10  & \cellcolor[rgb]{ .984,  .886,  .835}48.93\% & 69.97  & 95.42  & 61.36\% & \cellcolor[rgb]{ .984,  .886,  .835}69.65  & \cellcolor[rgb]{ .984,  .886,  .835}90.07  & \cellcolor[rgb]{ .984,  .886,  .835}110.38\% \\
\cmidrule{2-20}    \multirow{2}[2]{*}{\textbf{Mean}} & Global & 11.83  & 16.92  & 21.38\% & 5.94  & 9.39  & 15.10\% & 27.45  & 47.17  & 41.54\% & 30.96  & 51.87  & 22.51\% & 34.89  & 54.58  & 24.01\% & 28.12  & 47.01  & 32.45\% \\
          & Unobserved & 12.95  & 18.76  & 24.21\% & 6.39  & 9.90  & 16.89\% & 67.43  & 91.86  & 130.99\% & 71.33  & 97.98  & 49.82\% & 70.47  & 95.92  & 60.71\% & 71.73  & 92.07  & 113.88\% \\
    \midrule
    \midrule
    \multirow{2}[1]{*}{\textbf{TPB}} & Global & 12.63  & 18.00  & 22.72\% & 6.56  & 10.48  & 16.65\% & 29.46  & 51.06  & 44.22\% & 33.33  & 56.29  & 24.34\% & 37.18  & 58.77  & 25.13\% & 30.07  & 50.54  & 34.76\% \\
          & Unobserved & 13.83  & 19.96  & 25.72\% & 7.06  & 11.05  & 18.63\% & 72.36  & 99.43  & 139.42\% & 76.80  & 106.34  & 53.86\% & 75.10  & 103.28  & 63.55\% & 76.70  & 98.98  & 121.97\% \\
    \multirow{2}[0]{*}{\textbf{TransGTR}} & Global & 12.23  & 17.85  & 22.01\% & 6.36  & 10.39  & 16.13\% & 28.54  & 50.62  & 42.84\% & 32.29  & 55.80  & 23.58\% & 36.02  & 56.26  & 23.34\% & 29.13  & 50.11  & 33.68\% \\
          & Unobserved & 13.39  & 19.79  & 24.92\% & 6.84  & 10.96  & 18.05\% & 70.10  & 98.57  & 135.07\% & 74.41  & 105.41  & 52.18\% & 72.75  & 102.39  & 61.55\% & 74.31  & 98.14  & 118.18\% \\
    \multirow{2}[1]{*}{\textbf{STGP}} & Global & 11.74  & 17.19  & 21.13\% & 6.10  & 9.97  & 15.48\% & 27.41  & 48.74  & 41.13\% & 31.03  & 53.73  & 22.64\% & 36.58  & 56.11  & \cellcolor[rgb]{ .855,  .914,  .973}23.37\% & 27.97  & 48.25  & 32.33\% \\
          & Unobserved & 12.86  & 19.06  & 23.92\% & 6.56  & 10.52  & 17.32\% & 67.32  & 94.91  & 129.68\% & 71.50  & 101.50  & 50.10\% & \cellcolor[rgb]{ .984,  .886,  .835}69.85  & \cellcolor[rgb]{ .984,  .886,  .835}98.61  & \cellcolor[rgb]{ .984,  .886,  .835}59.10\% & 71.35  & 94.50  & 113.44\% \\
\cmidrule{2-20}    \multirow{2}[2]{*}{\textbf{Mean}} & Global & 12.20  & 17.68  & 21.95\% & 6.34  & 10.28  & 16.09\% & 28.47  & 50.14  & 42.73\% & 32.22  & 55.27  & 23.52\% & 36.59  & 57.05  & 23.95\% & 29.06  & 49.63  & 33.59\% \\
          & Unobserved & 13.36  & 19.60  & 24.85\% & 6.82  & 10.84  & 18.00\% & 69.93  & 97.64  & 134.72\% & 74.24  & 104.42  & 52.05\% & 72.57  & 101.43  & 61.40\% & 74.12  & 97.21  & 117.86\% \\
    \midrule
    \midrule
    \multirow{2}[2]{*}{\textbf{RoadDiff}} & Global & 12.31  & 17.62  & 21.69\% & 6.58  & 10.23  & 16.21\% & 28.12  & 47.74  & 41.64\% & 32.51  & 53.86  & 22.85\% & 36.88  & 56.89  & 25.06\% & 29.07  & 48.19  & 32.49\% \\
          & Unobserved & 13.48  & 19.54  & 24.56\% & 7.08  & 10.79  & 18.14\% & 69.07  & 92.96  & 131.29\% & 74.91  & 101.75  & 50.56\% & 74.49  & 99.98  & 63.37\% & 74.15  & 94.38  & 114.00\% \\
    \midrule
    \midrule
    \multirow{2}[2]{*}{\textbf{GenST}} & Global & \cellcolor[rgb]{ .651,  .788,  .925}\textbf{4.15 } & \cellcolor[rgb]{ .651,  .788,  .925}\textbf{8.15 } & \cellcolor[rgb]{ .651,  .788,  .925}\textbf{10.31\%} & \cellcolor[rgb]{ .651,  .788,  .925}\textbf{2.90 } & \cellcolor[rgb]{ .651,  .788,  .925}\textbf{6.35 } & \cellcolor[rgb]{ .651,  .788,  .925}\textbf{7.14\%} & \cellcolor[rgb]{ .651,  .788,  .925}\textbf{24.72 } & \cellcolor[rgb]{ .651,  .788,  .925}\textbf{38.42 } & \cellcolor[rgb]{ .651,  .788,  .925}\textbf{26.46\%} & \cellcolor[rgb]{ .651,  .788,  .925}\textbf{30.50 } & \cellcolor[rgb]{ .651,  .788,  .925}\textbf{48.50 } & \cellcolor[rgb]{ .651,  .788,  .925}\textbf{21.68\%} & 35.97  & 54.91  & \cellcolor[rgb]{ .651,  .788,  .925}\textbf{17.33\%} & \cellcolor[rgb]{ .651,  .788,  .925}\textbf{24.13 } & \cellcolor[rgb]{ .651,  .788,  .925}\textbf{36.94 } & \cellcolor[rgb]{ .651,  .788,  .925}\textbf{16.83\%} \\
          & Unobserved & \cellcolor[rgb]{ .945,  .663,  .514}\textbf{5.57 } & \cellcolor[rgb]{ .945,  .663,  .514}\textbf{9.89 } & \cellcolor[rgb]{ .945,  .663,  .514}\textbf{16.65\%} & \cellcolor[rgb]{ .945,  .663,  .514}\textbf{4.38 } & \cellcolor[rgb]{ .945,  .663,  .514}\textbf{8.89 } & \cellcolor[rgb]{ .945,  .663,  .514}\textbf{12.40\%} & \cellcolor[rgb]{ .945,  .663,  .514}\textbf{21.29 } & \cellcolor[rgb]{ .945,  .663,  .514}\textbf{32.39 } & \cellcolor[rgb]{ .945,  .663,  .514}\textbf{40.74\%} & \cellcolor[rgb]{ .945,  .663,  .514}\textbf{27.40 } & \cellcolor[rgb]{ .945,  .663,  .514}\textbf{41.41 } & \cellcolor[rgb]{ .945,  .663,  .514}\textbf{20.23\%} & \cellcolor[rgb]{ .945,  .663,  .514}\textbf{32.54 } & \cellcolor[rgb]{ .945,  .663,  .514}\textbf{47.79 } & \cellcolor[rgb]{ .945,  .663,  .514}\textbf{20.50\%} & \cellcolor[rgb]{ .945,  .663,  .514}\textbf{19.76 } & \cellcolor[rgb]{ .945,  .663,  .514}\textbf{28.81 } & \cellcolor[rgb]{ .945,  .663,  .514}\textbf{21.40\%} \\
    \bottomrule
    \end{tabular}%
    }
    \vspace{-0.4cm}
  \label{tab:main}%
\end{table*}%

We evaluated the performance of GenST against 16 baseline models under the FUNS setting and conducted ablation, efficiency, robustness, and cross-domain studies. Additional robustness and sensitivity analyses are provided in the \textbf{Appendix}.

\subsection{Experimental Settings}
\subsubsection{Datasets Description}
We use eight real-world benchmarks, including six traffic datasets, i.e., METR-LA and PeMS-Bay for traffic speed, and PeMS03, PeMS04, PeMS07, and PeMS08 for traffic flow, as well as two non-traffic datasets, KnowAir and Electricity, for cross-domain validation. To simulate FUNS, a fixed percentage of nodes is randomly selected as unobserved nodes ($V^-$), whose historical observations are removed during both training and inference. Detailed statistics are provided in \textbf{Appendix~\ref{Asec:Datasets}}.
\subsubsection{Baselines and Adaptation}We compared GenST with 16 baseline models across five distinct categories. Notably, methods capable only of historical data imputation without future forecasting were excluded. The categories include:
\textbf{Statistical Learning}: K-NN \cite{steinbach2009knn}, Kriging \cite{kleijnen2009kriging}, and HA \cite{campbell2008predicting}.
\textbf{End-to-End Deep Learning}: STGCN \cite{yu2018spatio}, GWNet \cite{wu2019graph}, AGCRN \cite{bai2020adaptive}, STGODE \cite{fang2021spatial}, GTS \cite{shang2021discrete}, STID \cite{shao2022spatial}, and MegaCRN \cite{jiang2023spatio}.
\textbf{Pre-trained Models}: STEP \cite{shao2022pre} and ST-LLM+~\cite{liu2025st}.
\textbf{Transfer Learning}: TPB \cite{liu2023cross}, TransGTR \cite{jin2023transferable}, and STGP \cite{hu2024prompt}.
\textbf{Graph Node Generation}: RoadDiff (the first model to address graph node generation) \cite{li2025fine}. Specific adaptations and fairness checks for the various baselines are detailed in \textbf{Appendix~\ref{Asec: baseline}}.
\subsubsection{Evaluation Metrics}
Following standard evaluation protocols, we utilized three core metrics to quantify prediction performance: Mean Absolute Error (MAE), Root Mean Square Error (RMSE), and Mean Absolute Percentage Error (MAPE), where lower values indicate better prediction accuracy and are detailed in \textbf{Appendix~\ref{Asec:Metric}}.

\subsubsection{Implementation Details}
All experiments were conducted on an Intel Xeon CPU Max 9462 (2.70GHz) server with six NVIDIA A100 80GB SXM GPUs. We split all datasets chronologically into training, validation, and test sets with a ratio of 7:1:2, and set both the historical window $T$ and prediction horizon $h$ to 12. Data are normalized by Z-Score. GenST uses LLaMA3.1-8B as the default backend LLM and $K=1000$ diffusion steps. All models are trained with Adam for up to 100 epochs with early stopping, an initial learning rate of $1 \times 10^{-3}$, cosine annealing, and batch size 64. All reported results are averaged over three random seeds. For baselines, we reproduce official implementations as closely as possible and adapt inputs for FUNS following \textbf{Appendix~\ref{Asec: baseline}}.

\vspace{-0.3cm}
\subsection{Main Experimental Analysis}
\textbf{Table~\ref{tab:main}} reports the performance of GenST and baseline models on the two core FUNS tasks: Global Network and Unobserved Node Forecasting. We also report category-wise averages to compare the suitability of different modeling paradigms.

GenST shows a clear advantage in the challenging unobserved-node prediction task. In terms of Unobserved MAE, GenST reduces errors by 56.0\% on METR-LA and 33.4\% on PeMS-Bay compared with the runner-up pre-trained model ST-LLM+, and the reduction further reaches 67.1\% on the more difficult PeMS03 flow dataset. Although GenST slightly trails ST-LLM+ in PeMS07 global forecasting, it still achieves the best performance on the corresponding unobserved task. This indicates that pre-trained models may still rely on transductive historical observations, whereas GenST’s inductive design better matches the zero-shot nature of FUNS.

The performance gap also varies across dataset types. On speed datasets, baselines show relatively small gaps between global and unobserved forecasting errors, such as the 1.35 MAE difference of STGCN on METR-LA. In contrast, the gap becomes much larger on flow datasets, where some baselines exceed 150\% Unobserved MAPE on PeMS03. This is because speed tends to be spatially smoother, while flow is highly heterogeneous across road functions and magnitudes. Without historical anchors, baselines struggle to infer absolute scales for unobserved nodes. GenST alleviates this issue through multi-modal semantic alignment, where LLM-derived descriptions provide functional and scale priors for zero-shot physical calibration. As a result, GenST effectively controls errors on flow datasets and establishes a stronger benchmark for FUNS.

\subsection{Impact of Unobserved Node Ratios \label{sec:4}}
\begin{figure}[t]
\vspace{-0.4cm}
    \centering
    \includegraphics[width=0.99\linewidth]{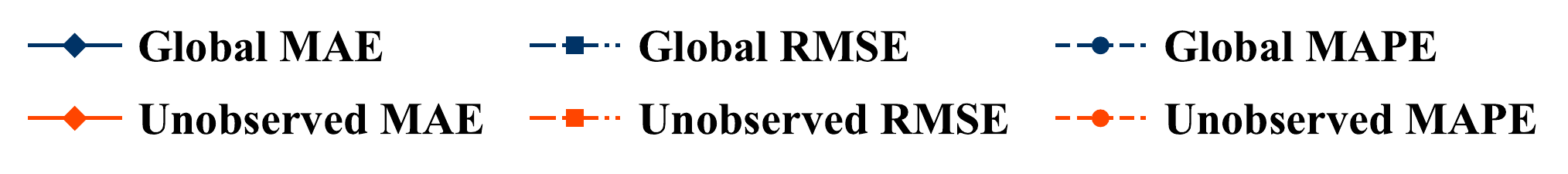}
    % \vspace{-0.5cm}
    \begin{subfigure}[b]{0.48\linewidth}
        \centering
        \includegraphics[width=\linewidth]{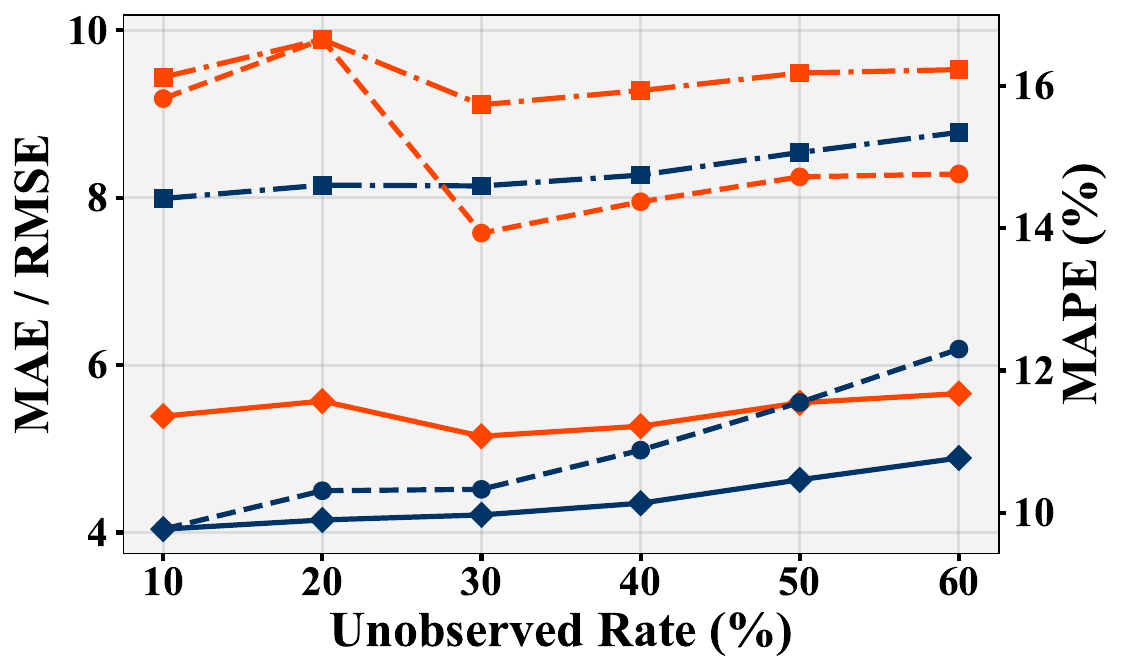}
        \vspace{-0.5cm}
        \caption{On METR-LA.}
    \end{subfigure}
    \hfill 
    \begin{subfigure}[b]{0.48\linewidth}
        \centering
        \includegraphics[width=\linewidth]{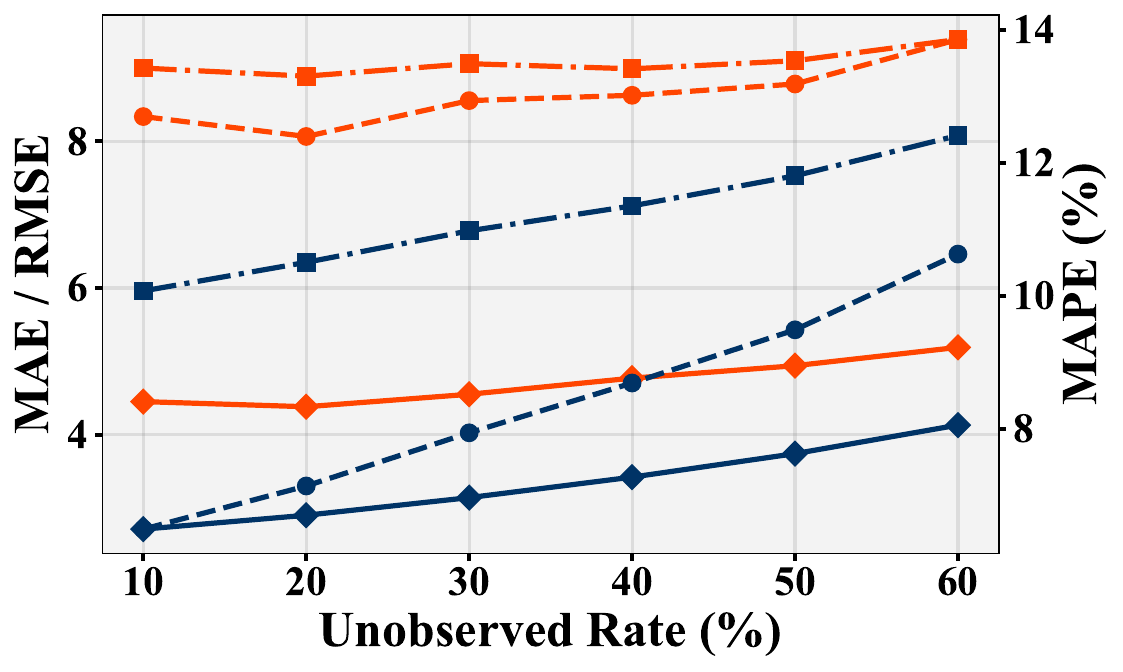}
        \vspace{-0.5cm}
        \caption{On PeMS-Bay.}
    \end{subfigure}
    \vspace{0.4cm} 
    \begin{subfigure}[b]{0.48\linewidth}
        \centering
        \includegraphics[width=\linewidth]{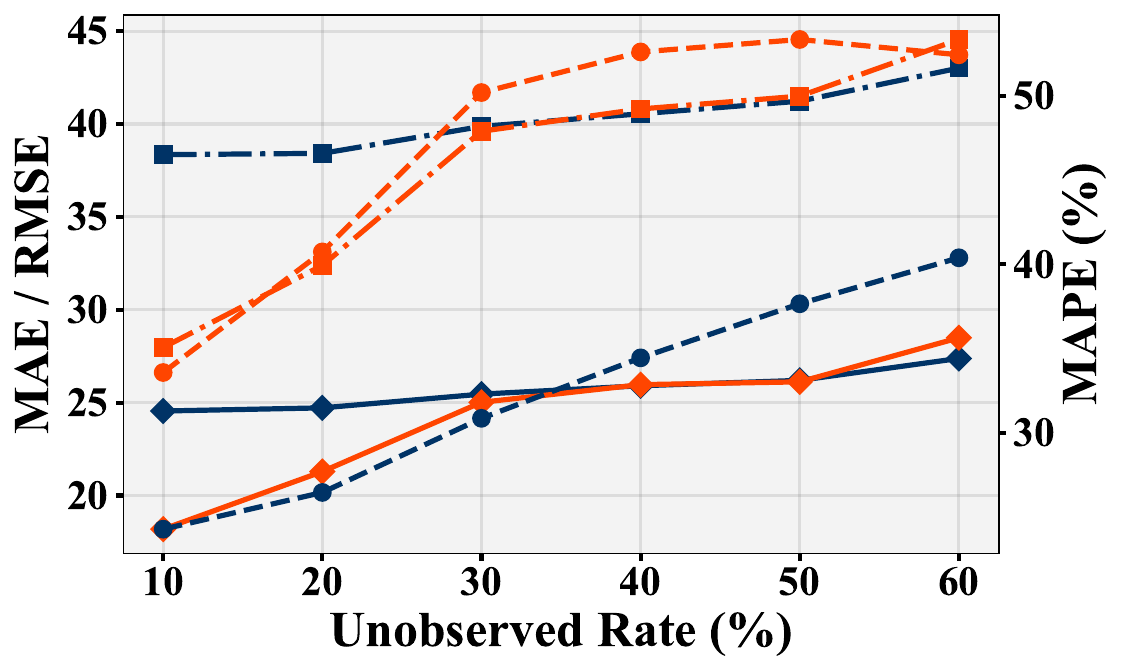}
        \vspace{-0.5cm}
        \caption{On PeMS03.}
    \end{subfigure}
    \hfill 
    \begin{subfigure}[b]{0.48\linewidth}
        \centering
        \includegraphics[width=\linewidth]{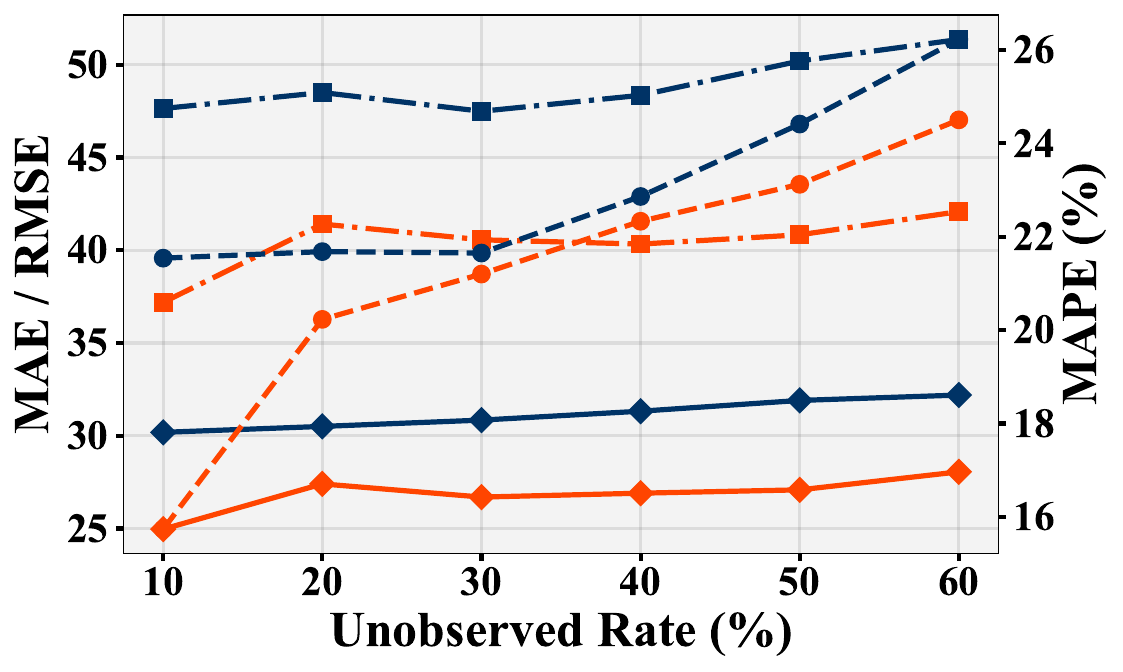}
        \vspace{-0.5cm}
        \caption{On PeMS04.}
    \end{subfigure}
    \vspace{-0.8cm}
    \caption{Varying unobserved ratios robustness evaluation.}
    \vspace{-0.3cm}
    \label{fig:missing}
\end{figure}

\textbf{Figure~\ref{fig:missing}} illustrates GenST's performance across four datasets with unobserved node ratios ranging from 10\% to 60\%. As the proportion of unobserved nodes increases, the reduction in available information generally leads to a monotonic increase in global prediction error. In contrast, GenST remains robust on unobserved-node prediction by leveraging stable priors from multi-modal semantic alignment. For example, on PeMS-Bay and PeMS04, the Unobserved MAE stays nearly flat even when the unobserved ratio increases from 10\% to 60\%, with PeMS04 rising only from approximately 25 to 28. This indicates that GenST mitigates over-reliance on dense local observations by effectively exploiting semantic and coordinate information for zero-shot inference. We also observe a non-monotonic fluctuation on METR-LA, where the errors at 10\% and 20\% are higher than that at 30\%. This is mainly due to sample bias: when the missing ratio is low, the small unobserved set $V^-$ may randomly include hard nodes with complex topology or high noise, whereas larger ratios make $V^-$ more representative of the full graph and thus yield more stable evaluation.

% Figure 4 illustrates GenST's performance across four datasets with unobserved node ratios from 10\% to 60\%. While global prediction error (blue curve) rises monotonically as available information decreases, GenST’s unobserved node prediction (orange curve) remains remarkably robust due to stable multi-modal semantic priors. On PeMS-Bay and PeMS04, Unobserved MAE stays nearly flat (e.g., rising only from 25 to 28 in PeMS04) despite the ratio surge. This confirms that GenST effectively utilizes semantics and coordinates for zero-shot inference, reducing over-reliance on dense local observations. Notably, METR-LA exhibits non-monotonic fluctuations where errors at 10\%–20\% exceed those at 30\%. This "sample bias" occurs because small $V^-$ sets at low ratios may randomly target "hard nodes" with complex topologies or high noise. At 30\% and above, the $V^-$ distribution becomes more representative of the full graph, leading to more stable and representative outcomes.

\begin{figure}[t]
\vspace{-0.2cm}
    \begin{subfigure}[b]{0.48\linewidth}
        \centering
        \includegraphics[width=\linewidth]{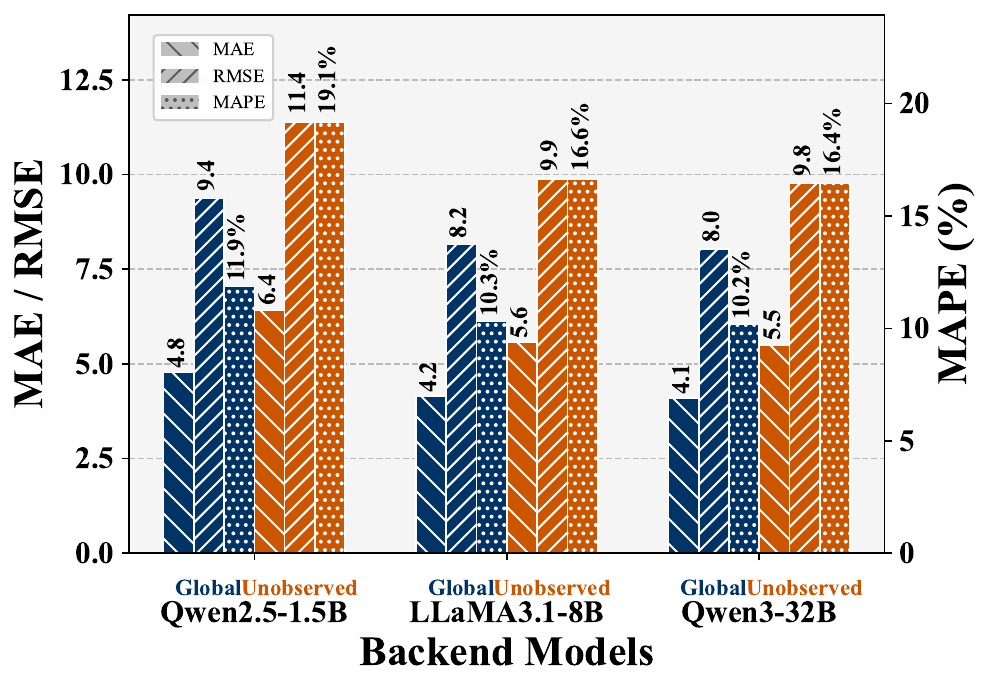}
        \vspace{-0.5cm}
        \caption{Accuracy on METR-LA.}
    \end{subfigure}
    \hfill 
    \begin{subfigure}[b]{0.48\linewidth}
        \centering
        \includegraphics[width=\linewidth]{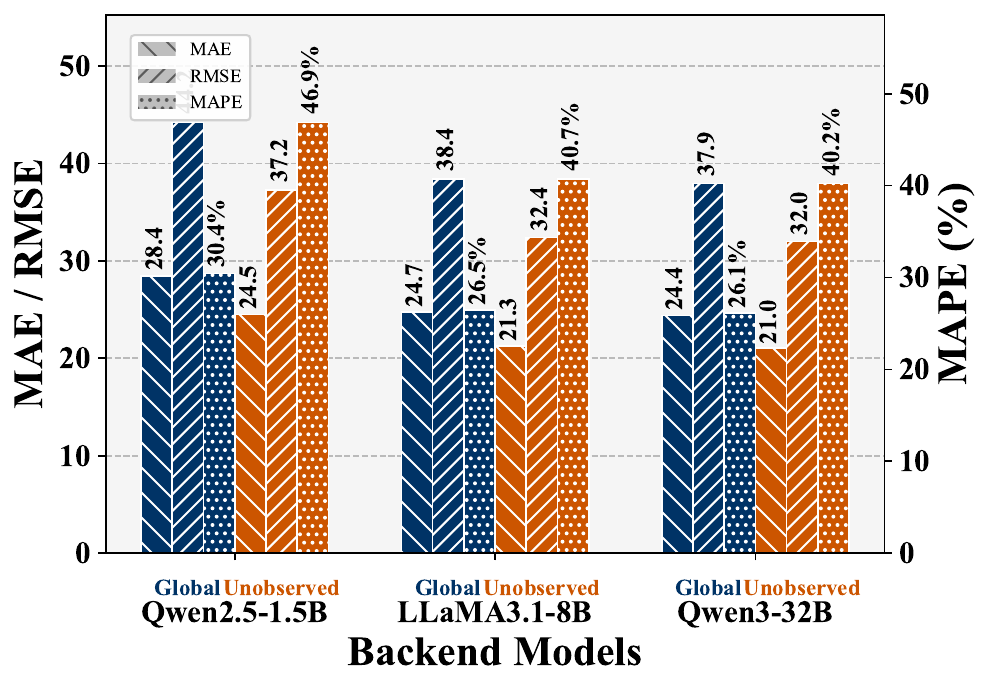}
        \vspace{-0.5cm}
        \caption{Accuracy on PeMS03.}
    \end{subfigure}
    \vspace{0.2cm} 
    \begin{subfigure}[b]{0.48\linewidth}
        \centering
        \includegraphics[width=\linewidth]{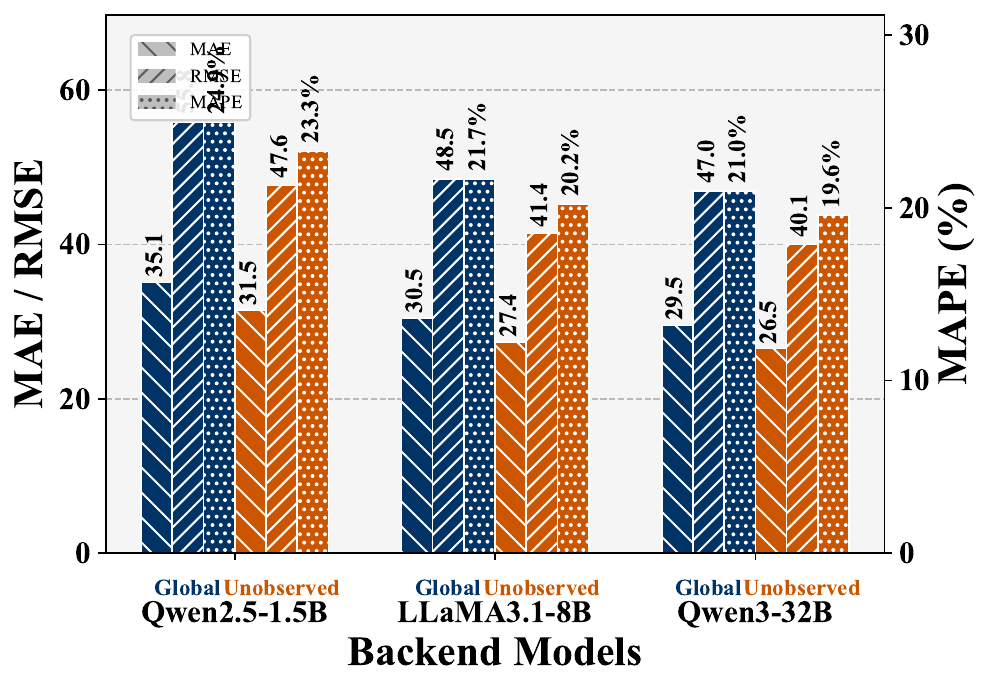}
        \vspace{-0.5cm}
        \caption{Accuracy on PeMS04.}
    \end{subfigure}
    \hfill 
    \begin{subfigure}[b]{0.48\linewidth}
        \centering
        \includegraphics[width=\linewidth]{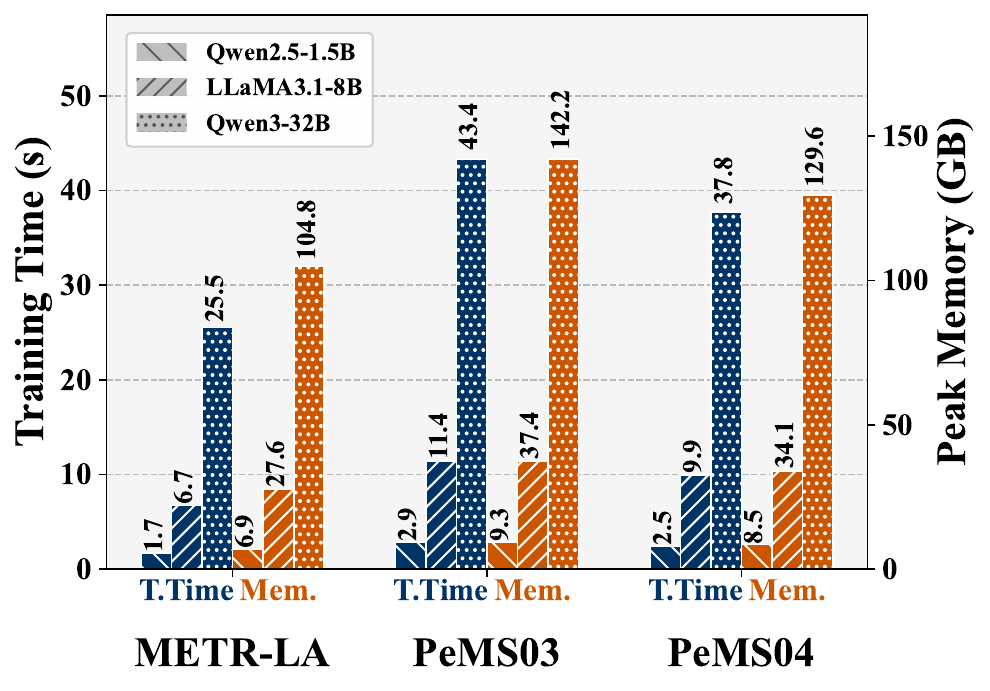}
        \vspace{-0.5cm}
        \caption{Training efficiency.}
    \end{subfigure}
   \vspace{-0.6cm}
    \caption{Backend LLM comparison.}
    \vspace{-0.6cm}
    \label{fig:backend}
\end{figure}

\subsection{Study of Different Backend LLMs}
To evaluate the impact of LLM scale and reasoning capabilities on GenST, we compared three backends: Qwen2.5-1.5B (lightweight), Llama3.1-8B (medium), and Qwen3-32B (large). \textbf{Figures~\ref{fig:backend}(a)–(c)} illustrate prediction accuracy across datasets, while \textbf{Figures~\ref{fig:backend}(d)} details resource consumption in terms of training time and memory.

Results demonstrate a positive correlation between model scale and performance. While Qwen2.5-1.5B exhibits higher errors due to limited reasoning, scaling to 8B and 32B significantly improves accuracy, confirming that superior reasoning enhances inductive bias for zero-shot inference. However, marginal returns diminish; the gain from 8B to 32B is notably smaller than from 1.5B to 8B. Efficiency-wise, Qwen3-32B achieves peak accuracy but is extremely costly—training 3.3x longer than 8B, with peak VRAM reaching 129.6 GB. Conversely, Llama3.1-8B provides near-SOTA performance with low memory overhead and fast training, representing the optimal trade-off between accuracy and efficiency.

\subsection{Ablation Study}
To validate the contributions of GenST's core components, we design four ablation variants by removing the Semantic Alignment Module (w/o S.Alig.), Spatio-Temporal VAE (w/o ST-VAE), Cross-Attention (w/o C.Att), and Spatial Context (w/o Sp. Con.). \textbf{Table~\ref{tab:ablation}} reports the impact of each removal on prediction accuracy. The Semantic Alignment Module is the most critical component, as removing it causes the largest performance degradation. For example, on PeMS03, Unobserved MAE increases by 9.33, and MAPE worsens by 17.54\% without semantic alignment. This confirms that LLM-provided open-world semantics are essential for zero-shot inference, as they help infer physical scales for nodes without historical observations and reduce severe scale confusion. Among the remaining components, removing ST-VAE leads to the second largest degradation. On PeMS03, the MAE increase caused by removing ST-VAE (+7.12) is notably larger than that caused by removing Spatial Context (+0.11), indicating the importance of latent manifold learning for robust generation. Removing Cross-Attention also consistently worsens performance, showing that conditional knowledge injection is necessary for effectively using multi-modal guidance. Spatial Context has the smallest contribution, e.g., +0.28 MAE on METR-LA and only +0.11 on PeMS03 (within seed variance), suggesting that after semantics capture macro-level patterns, coordinates and neighborhood context mainly provide fine-grained spatial calibration.

\begin{table}[t]
  \centering
  \vspace{-0.5cm}
  \caption{Components ablation study.}
  \vspace{-0.4cm}
  \resizebox{\linewidth}{!}{
    \setlength{\tabcolsep}{1.5pt}
    \begin{tabular}{cccccccc}
    \toprule
    \multirow{2}[3]{*}{\textbf{Variants}} & \textbf{Datasets} & \multicolumn{3}{c}{\textbf{METR-LA}} & \multicolumn{3}{c}{\textbf{PeMS-Bay}} \\
    \cmidrule(lr){3-5} \cmidrule(lr){6-8}
          & \textbf{Metrics} & \textbf{MAE} & \textbf{RMSE} & \textbf{MAPE} & \textbf{MAE} & \textbf{RMSE} & \textbf{MAPE} \\
    \midrule
    
    \textbf{w/o} & \textbf{Global} & 5.23 \textcolor{myblue}{(\(\uparrow\)1.08)} & 9.30 \textcolor{myblue}{(\(\uparrow\)1.15)} & 12.65\% \textcolor{myblue}{(\(\uparrow\)2.34\%)} & 3.90 \textcolor{myblue}{(\(\uparrow\)1.00)} & 7.26 \textcolor{myblue}{(\(\uparrow\)0.91)} & 8.20\% \textcolor{myblue}{(\(\uparrow\)1.06\%)} \\
    \textbf{ S.Alig.} & \textbf{Unob.} & 7.68 \textcolor{myorange}{(\(\uparrow\)2.11)} & 13.22 \textcolor{myorange}{(\(\uparrow\)3.33)} & 27.01\% \textcolor{myorange}{(\(\uparrow\)10.36\%)} & 6.23 \textcolor{myorange}{(\(\uparrow\)1.85)} & 12.20 \textcolor{myorange}{(\(\uparrow\)3.31)} & 21.07\% \textcolor{myorange}{(\(\uparrow\)8.67\%)} \\
    
    \textbf{w/o } & \textbf{Global} & 4.73 \textcolor{myblue}{(\(\uparrow\)0.58)} & 8.81 \textcolor{myblue}{(\(\uparrow\)0.66)} & 11.58\% \textcolor{myblue}{(\(\uparrow\)1.27\%)} & 3.55 \textcolor{myblue}{(\(\uparrow\)0.65)} & 6.87 \textcolor{myblue}{(\(\uparrow\)0.52)} & 7.67\% \textcolor{myblue}{(\(\uparrow\)0.53\%)} \\
    \textbf{ST-VAE} & \textbf{Unob.} & 6.26 \textcolor{myorange}{(\(\uparrow\)0.69)} & 10.56 \textcolor{myorange}{(\(\uparrow\)0.67)} & 17.94\% \textcolor{myorange}{(\(\uparrow\)1.29\%)} & 5.63 \textcolor{myorange}{(\(\uparrow\)1.25)} & 9.52 \textcolor{myorange}{(\(\uparrow\)0.63)} & 13.29\% \textcolor{myorange}{(\(\uparrow\)0.89\%)} \\
 
    \textbf{w/o } & \textbf{Global} & 4.61 \textcolor{myblue}{(\(\uparrow\)0.46)} & 8.60 \textcolor{myblue}{(\(\uparrow\)0.45)} & 11.03\% \textcolor{myblue}{(\(\uparrow\)0.72\%)} & 3.40 \textcolor{myblue}{(\(\uparrow\)0.50)} & 6.66 \textcolor{myblue}{(\(\uparrow\)0.31)} & 7.54\% \textcolor{myblue}{(\(\uparrow\)0.40\%)} \\
    \textbf{C.Att} & \textbf{Unob.} & 6.08 \textcolor{myorange}{(\(\uparrow\)0.51)} & 10.29 \textcolor{myorange}{(\(\uparrow\)0.40)} & 17.39\% \textcolor{myorange}{(\(\uparrow\)0.74\%)} & 5.24 \textcolor{myorange}{(\(\uparrow\)0.86)} & 9.25 \textcolor{myorange}{(\(\uparrow\)0.36)} & 12.90\% \textcolor{myorange}{(\(\uparrow\)0.50\%)} \\
 
    \textbf{w/o } & \textbf{Global} & 4.36 \textcolor{myblue}{(\(\uparrow\)0.21)} & 8.16 \textcolor{myblue}{(\(\uparrow\)0.01)} & 10.52\% \textcolor{myblue}{(\(\uparrow\)0.21\%)} & 3.24 \textcolor{myblue}{(\(\uparrow\)0.34)} & 6.41 \textcolor{myblue}{(\(\uparrow\)0.06)} & 7.23\% \textcolor{myblue}{(\(\uparrow\)0.09\%)} \\
    \textbf{Sp. Con.} & \textbf{Unob.} & 5.85 \textcolor{myorange}{(\(\uparrow\)0.28)} & 9.92 \textcolor{myorange}{(\(\uparrow\)0.03)} & 16.80\% \textcolor{myorange}{(\(\uparrow\)0.15\%)} & 5.12 \textcolor{myorange}{(\(\uparrow\)0.74)} & 9.04 \textcolor{myorange}{(\(\uparrow\)0.15)} & 12.53\% \textcolor{myorange}{(\(\uparrow\)0.13\%)} \\
    
    \midrule
    
    \multirow{2}[4]{*}{\textbf{Variants}} & \textbf{Datasets} & \multicolumn{3}{c}{\textbf{PeMS03}} & \multicolumn{3}{c}{\textbf{PeMS04}} \\
    \cmidrule(lr){3-5} \cmidrule(lr){6-8}
          & \textbf{Metrics} & \textbf{MAE} & \textbf{RMSE} & \textbf{MAPE} & \textbf{MAE} & \textbf{RMSE} & \textbf{MAPE} \\
    \midrule
 
    \textbf{w/o} & \textbf{Global} & 29.42 \textcolor{myblue}{(\(\uparrow\)4.70)} & 44.68 \textcolor{myblue}{(\(\uparrow\)6.26)} & 30.31\% \textcolor{myblue}{(\(\uparrow\)3.85\%)} & 34.01 \textcolor{myblue}{(\(\uparrow\)3.51)} & 55.97 \textcolor{myblue}{(\(\uparrow\)7.47)} & 24.90\% \textcolor{myblue}{(\(\uparrow\)3.22\%)} \\
    \textbf{ S.Alig.} & \textbf{Unob.} & 30.62 \textcolor{myorange}{(\(\uparrow\)9.33)} & 42.92 \textcolor{myorange}{(\(\uparrow\)10.53)} & 58.28\% \textcolor{myorange}{(\(\uparrow\)17.54\%)} & 37.12 \textcolor{myorange}{(\(\uparrow\)9.72)} & 57.70 \textcolor{myorange}{(\(\uparrow\)16.29)} & 30.96\% \textcolor{myorange}{(\(\uparrow\)10.73\%)} \\
    \textbf{w/o } & \textbf{Global} & 27.58 \textcolor{myblue}{(\(\uparrow\)2.86)} & 41.23 \textcolor{myblue}{(\(\uparrow\)2.81)} & 28.59\% \textcolor{myblue}{(\(\uparrow\)2.13\%)} & 32.45 \textcolor{myblue}{(\(\uparrow\)1.95)} & 52.32 \textcolor{myblue}{(\(\uparrow\)3.82)} & 23.51\% \textcolor{myblue}{(\(\uparrow\)1.83\%)} \\
    \textbf{ST-VAE} & \textbf{Unob.} & 28.41 \textcolor{myorange}{(\(\uparrow\)7.12)} & 37.49 \textcolor{myorange}{(\(\uparrow\)5.10)} & 43.78\% \textcolor{myorange}{(\(\uparrow\)3.04\%)} & 31.94 \textcolor{myorange}{(\(\uparrow\)4.54)} & 44.96 \textcolor{myorange}{(\(\uparrow\)3.55)} & 21.82\% \textcolor{myorange}{(\(\uparrow\)1.59\%)} \\
    \textbf{w/o } & \textbf{Global} & 26.17 \textcolor{myblue}{(\(\uparrow\)1.45)} & 40.11 \textcolor{myblue}{(\(\uparrow\)1.69)} & 27.76\% \textcolor{myblue}{(\(\uparrow\)1.30\%)} & 31.70 \textcolor{myblue}{(\(\uparrow\)1.20)} & 50.37 \textcolor{myblue}{(\(\uparrow\)1.87)} & 22.45\% \textcolor{myblue}{(\(\uparrow\)0.77\%)} \\
    \textbf{C.Att} & \textbf{Unob.} & 25.17 \textcolor{myorange}{(\(\uparrow\)3.88)} & 34.77 \textcolor{myorange}{(\(\uparrow\)2.38)} & 42.26\% \textcolor{myorange}{(\(\uparrow\)1.52\%)} & 31.42 \textcolor{myorange}{(\(\uparrow\)4.02)} & 43.99 \textcolor{myorange}{(\(\uparrow\)2.58)} & 21.23\% \textcolor{myorange}{(\(\uparrow\)1.00\%)} \\
    \textbf{w/o } & \textbf{Global} & 25.10 \textcolor{myblue}{(\(\uparrow\)0.38)} & 38.68 \textcolor{myblue}{(\(\uparrow\)0.26)} & 26.81\% \textcolor{myblue}{(\(\uparrow\)0.35\%)} & 31.16 \textcolor{myblue}{(\(\uparrow\)0.66)} & 48.64 \textcolor{myblue}{(\(\uparrow\)0.14)} & 21.81\% \textcolor{myblue}{(\(\uparrow\)0.13\%)} \\
    \textbf{Sp. Con.} & \textbf{Unob.} & 21.40 \textcolor{myorange}{(\(\uparrow\)0.11)} & 32.74 \textcolor{myorange}{(\(\uparrow\)0.35)} & 40.86\% \textcolor{myorange}{(\(\uparrow\)0.12\%)} & 28.93 \textcolor{myorange}{(\(\uparrow\)1.53)} & 41.66 \textcolor{myorange}{(\(\uparrow\)0.25)} & 20.24\% \textcolor{myorange}{(\(\uparrow\)0.01\%)} \\
    \bottomrule
    \end{tabular}%
    }
  \label{tab:ablation}%
  \vspace{-0.4cm}
\end{table}

\subsection{Efficiency Analysis}
\textbf{Table~\ref{tab:cost}} reports the training and inference cost of GenST under different backbone and sampler configurations. Thanks to LoRA, the number of trainable parameters remains 36.98M regardless of node scale, validating the parameter efficiency of our inductive design. Under the default LLaMA3.1-8B with DDPM setting, training VRAM on PeMS04 reaches 34.09 GB, reflecting the offline cost of the 8B backbone and 1000-step diffusion. GenST targets high-accuracy zero-shot FUNS, while accelerated sampling and lightweight LLM backbones provide controllable accuracy-efficiency trade-offs.

DDIM and DPM-Solver++ serve as drop-in replacements for DDPM without retraining. On PeMS04, DPM-Solver++ reduces sampling from 1000 to 20 steps and cuts latency from 3.17s to 0.78s, achieving a 4.1$\times$ speedup with only a 12.6\% Unobserved MAE cost. Compared with non-generative inductive baselines on the same hardware, GenST still maintains sub-second latency while reducing Unobserved MAE by 57\% over the strongest baseline. Under tighter resource budgets, replacing the backend with Qwen2.5-1.5B further reduces training VRAM to 8.54 GB, about one quarter of the 8B setting, with proportionally shorter training time at roughly 15\% accuracy cost. These results show that GenST is training-heavy but inference-efficient, and can flexibly trade accuracy for resource efficiency across deployment scenarios.

\begin{table}[t]
  \centering
  \vspace{-0.3cm}
  \caption{Training and inference cost.}
  \vspace{-0.3cm}
  \resizebox{0.95\linewidth}{!}{
  \setlength{\tabcolsep}{2pt}
     \begin{tabular}{cc|cc|c|ccc}
    \toprule
    \multirow{2}[4]{*}{\textbf{Dataset}} & \multirow{2}[4]{*}{\textbf{Backend}} & \multicolumn{2}{c|}{\textbf{Training}} & \multirow{2}[4]{*}{\textbf{Sampler}} & \multicolumn{3}{c}{\textbf{Inference}} \\
\cmidrule{3-4}\cmidrule{6-8}          &       & \textbf{T. Time} & \textbf{T.Mem. (Peak)} &       & \textbf{In. Time} & \textbf{In.Mem. (Peak)} & \textbf{Unobs. MAE} \\
    \midrule
    \multirow{4}[4]{*}{\textbf{METR-LA}} & \multirow{3}[2]{*}{\textbf{LLaMA3.1-8B}} & \multirow{3}[2]{*}{6.72s} & \multirow{3}[2]{*}{27.59G} & DDPM (K=1000) & 2.15s & 17.49G & 5.57 \\
          &       &       &       & DDIM (K=50) & 0.73s & 17.49G & 6.33 \\
          &       &       &       & DPM-Solver++ (K=20) & 0.53s & 17.49G & 6.27 \\
\cmidrule{2-8}          & \textbf{Qwen2.5-1.5B} & 1.69s & 6.92G & DDPM (K=1000) & 0.49s & 3.20G & 6.41 \\
    \midrule
    \midrule
    \multirow{4}[4]{*}{\textbf{PeMS04}} & \multirow{3}[2]{*}{\textbf{LLaMA3.1-8B}} & \multirow{3}[2]{*}{9.93s} & \multirow{3}[2]{*}{34.09G} & DDPM (K=1000) & 3.17s & 19.12G & 27.40 \\
          &       &       &       & DDIM (K=50) & 1.08s & 19.12G & 31.15 \\
          &       &       &       & DPM-Solver++ (K=20) & 0.78s & 19.12G & 30.85 \\
\cmidrule{2-8}          & \textbf{Qwen2.5-1.5B} & 2.52s & 8.54G & DDPM (K=1000) & 0.80s & 4.01G & 31.51 \\
    \bottomrule
    \end{tabular}%
    }
\vspace{-0.7cm}
  \label{tab:cost}%
\end{table}%

\subsection{Cross-Domain Generalization}
\label{sec:cross_domain}

Although traffic is the predominant scenario for spatio-temporal graph modeling~\cite{lyu2025autostf,gao2024spatial}, we further test whether GenST transfers beyond traffic by applying the same architecture to KnowAir and Electricity. Semantic construction, adjacency processing, and the $V^-$ placeholder strategy are detailed in \textbf{Appendix~\ref{Asec:Datasets}}.

\textbf{Table~\ref{tab:cross_domain}} compares GenST with representative baselines from statistical, end-to-end, pre-trained, and generative paradigms. GenST achieves the best Unobserved performance on both datasets, substantially outperforming ST-LLM+, the strongest non-generative baseline. First, the Global--Unobserved gap remains small after domain shift, consistent with traffic datasets, indicating that GenST's inductive zero-shot capability transfers without retraining. Second, Electricity is a weak-metadata setting with anonymized clients, no native coordinates, and no POI-style semantics; unobserved nodes only receive generic semantic placeholders and structural context from the released adjacency. GenST still achieves the best unobserved-node performance, suggesting that the generative framework remains effective even with incomplete semantic annotations.

\section{Related Work}
\subsection{Spatio-temporal Forecasting}
Spatio-temporal forecasting has transitioned from statistical methods like HA~\cite{campbell2008predicting} and ARIMA~\cite{box2015time} to modern deep learning paradigms. While CNNs initially handled grid-based data~\cite{shi2015convolutional,zhang2017deep}, GNNs such as STGCN \cite{yu2018spatio}, GWNet \cite{wu2019graph}, and AGCRN \cite{bai2020adaptive} have become the standard for modeling non-Euclidean spatial dependencies. To improve generalization, pre-training frameworks like STEP \cite{shao2022pre} and ST-LLM+~\cite{liu2025st} learn universal patterns from large-scale datasets, yet they falter when sensor data is entirely missing.

Although transfer learning methods (TransGTR \cite{jin2023transferable}, TPB \cite{liu2023cross}) attempt to bridge this gap via domain adaptation or meta-learning, most existing models remain transductive. Their reliance on historical sequences or learnable node embeddings for identity presents a fundamental barrier to FUNS. Without history or pre-learned embeddings, these models cannot infer patterns for unseen nodes. While FUNS-N explored this problem \cite{roth2022forecasting}, it lacks mathematical rigor, extensive baseline comparisons, and open-source code, limiting its reproducibility. Consequently, the field lacks a systematic, inductive, and semantic-driven framework validated against broad baselines to address the FUNS task effectively.

\begin{table}[t]
  \centering
  \vspace{-0.2cm}
  \caption{Cross-domain generalization.}
  \vspace{-0.3cm}
  \resizebox{\linewidth}{!}{
  \setlength{\tabcolsep}{2pt}
      \begin{tabular}{cccccccccccccc}
    \toprule
    \multicolumn{2}{c}{\textbf{Models}} & \multicolumn{2}{c}{\textbf{K-NN}} & \multicolumn{2}{c}{\textbf{GWNet}} & \multicolumn{2}{c}{\textbf{AGCRN}} & \multicolumn{2}{c}{\textbf{ST-LLM+}} & \multicolumn{2}{c}{\textbf{RoadDiff}} & \multicolumn{2}{c}{\textbf{GenST}} \\
    \midrule
    \textbf{Datasets} & \textbf{Tasks} & \textbf{Global} & \textbf{Unob.} & \textbf{Global} & \textbf{Unob.} & \textbf{Global} & \textbf{Unob.} & \textbf{Global} & \textbf{Unob.} & \textbf{Global} & \textbf{Unob.} & \textbf{Global} & \textbf{Unob.} \\
    \midrule
    \multirow{3}[2]{*}{\textbf{KownAir}} & \textbf{MAE} & 28.07 & 47.99 & 20.72 & 42.16 & 22.42 & 45.03 & \cellcolor[rgb]{ .855,  .914,  .973}20.25  & \cellcolor[rgb]{ .984,  .886,  .835}39.22  & 20.84 & 41.26 & \cellcolor[rgb]{ .651,  .788,  .925}\textbf{18.56 } & \cellcolor[rgb]{ .945,  .663,  .514}\textbf{19.78 } \\
          & \textbf{RMSE} & 43.13 & 68.86 & 32.63 & 53.20 & 34.85 & 56.18 & \cellcolor[rgb]{ .855,  .914,  .973}32.49  & \cellcolor[rgb]{ .984,  .886,  .835}51.17  & 33.14 & 52.49 & \cellcolor[rgb]{ .651,  .788,  .925}\textbf{25.32 } & \cellcolor[rgb]{ .945,  .663,  .514}\textbf{30.14 } \\
          & \textbf{MAPE} & 34.32\% & 79.92\% & 29.21\% & 69.38\% & 32.20\% & 76.17\% & \cellcolor[rgb]{ .855,  .914,  .973}28.70\% & \cellcolor[rgb]{ .984,  .886,  .835}64.33\% & 29.54\% & 67.64\% & \cellcolor[rgb]{ .651,  .788,  .925}\textbf{18.82\%} & \cellcolor[rgb]{ .945,  .663,  .514}\textbf{22.53\%} \\
\cmidrule{2-14}    \multirow{3}[2]{*}{\textbf{Electricity}} & \textbf{MAE} & 3.13  & 5.68  & 2.58  & 5.78  & 2.72  & 6.41  & \cellcolor[rgb]{ .855,  .914,  .973}2.11  & \cellcolor[rgb]{ .984,  .886,  .835}4.66  & 2.26  & 5.12  & \cellcolor[rgb]{ .651,  .788,  .925}\textbf{1.87 } & \cellcolor[rgb]{ .945,  .663,  .514}\textbf{2.18 } \\
          & \textbf{RMSE} & 25.68  & 38.91  & 22.04  & 42.13  & 22.47  & 42.63  & \cellcolor[rgb]{ .855,  .914,  .973}17.50  & \cellcolor[rgb]{ .984,  .886,  .835}31.79  & 18.76  & 34.64  & \cellcolor[rgb]{ .651,  .788,  .925}\textbf{15.72 } & \cellcolor[rgb]{ .945,  .663,  .514}\textbf{19.37 } \\
          & \textbf{MAPE} & 24.46\% & 56.38\% & 16.84\% & 46.32\% & 21.42\% & 60.39\% & \cellcolor[rgb]{ .855,  .914,  .973}14.45\% & \cellcolor[rgb]{ .984,  .886,  .835}38.30\% & 16.12\% & 41.98\% & \cellcolor[rgb]{ .651,  .788,  .925}\textbf{13.68\%} & \cellcolor[rgb]{ .945,  .663,  .514}\textbf{19.12\%} \\
    \bottomrule
    \end{tabular}%
    }
  \label{tab:cross_domain}%
  \vspace{-0.4cm}
\end{table}%
\subsection{Graph-based Generation}
Graph generation has achieved success in bioinformatics and program synthesis using VAEs and Normalizing Flows, with models like MoFlow \cite{zang2020moflow}, GraphNVP \cite{madhawa2019graphnvp}, and RefineGNN synthesizing molecular or protein structures \cite{jiniterative}. However, these methods are tailored to specific domain constraints, such as chemical valency or syntax rules, making them difficult to repurpose for dynamic spatio-temporal traffic states.

In the spatio-temporal domain, generative research primarily focuses on data imputation (e.g., PriSTI \cite{liu2023pristi}, DiffSTG \cite{wen2023diffstg}) rather than node generation, reconstructing missing values conditioned on partial observations. RoadDiff is currently the only work attempting lane node generation by using a diffusion process based on road data \cite{li2025fine}. Nevertheless, RoadDiff is heavily constrained by explicit traffic flow theory and road engineering features, which limit its transferability to general spatio-temporal graphs or datasets lacking such physical attributes. This underscores the urgent need for a general, semantic-guided generative framework capable of inductive zero-shot inference.

% \vspace{-0.1cm}
\section{Conclusion and Future Work}
In this paper, we rigorously define the FUNS problem, decomposing it into complementary global and unobserved node forecasting tasks. We introduce GenST, which reformulates spatio-temporal forecasting as a conditional graph node generation task. By integrating LLM-extracted semantics with ST-VAE and latent diffusion, GenST establishes a fully inductive prediction paradigm for nodes without historical observations. Extensive experiments on six traffic benchmarks and two non-traffic datasets demonstrate that GenST consistently outperforms SOTA baselines in zero-shot unobserved-node forecasting, while additional analyses validate its robustness under sparse observations and accelerated sampling.
Despite its strong performance, GenST still involves relatively high offline training costs due to the LLM backbone and iterative diffusion sampling. Future work will further improve efficiency through knowledge distillation and model quantization, and extend GenST to irregular multivariate time series, grid-based spatio-temporal data, and larger-scale industrial sensing systems.
% \begin{acks}
% \end{acks}

% \clearpage
\bibliographystyle{ACM-Reference-Format}
\bibliography{sample-base}

@inproceedings{roth2022forecasting,
  title={Forecasting unobserved node states with spatio-temporal graph neural networks},
  author={Roth, Andreas and Liebig, Thomas},
  booktitle={2022 IEEE International Conference on Data Mining Workshops (ICDMW)},
  pages={740--747},
  year={2022},
  organization={IEEE}
}

@article{campbell2008predicting,
  title={Predicting excess stock returns out of sample: Can anything beat the historical average?},
  author={Campbell, John Y and Thompson, Samuel B},
  journal={The Review of Financial Studies},
  volume={21},
  number={4},
  pages={1509--1531},
  year={2008},
  publisher={Society for Financial Studies}
}

@book{box2015time,
  title={Time series analysis: forecasting and control},
  author={Box, George EP and Jenkins, Gwilym M and Reinsel, Gregory C and Ljung, Greta M},
  year={2015},
  publisher={John Wiley \& Sons}
}

@article{shi2015convolutional,
  title={Convolutional LSTM network: A machine learning approach for precipitation nowcasting},
  author={Shi, Xingjian and Chen, Zhourong and Wang, Hao and Yeung, Dit-Yan and Wong, Wai-Kin and Woo, Wang-chun},
  journal={Advances in neural information processing systems},
  volume={28},
  year={2015}
}

@inproceedings{zhang2017deep,
  title={Deep spatio-temporal residual networks for citywide crowd flows prediction},
  author={Zhang, Junbo and Zheng, Yu and Qi, Dekang},
  booktitle={Proceedings of the AAAI conference on artificial intelligence},
  volume={31},
  number={1},
  year={2017}
}

@article{liu2025st,
  title={ST-LLM+: Graph Enhanced Spatio-Temporal Large Language Models for Traffic Prediction},
  author={Liu, Chenxi and Hettige, Kethmi Hirushini and Xu, Qianxiong and Long, Cheng and Xiang, Shili and Cong, Gao and Li, Ziyue and Zhao, Rui},
  journal={IEEE Transactions on Knowledge and Data Engineering},
  year={2025},
  publisher={IEEE}
}

@inproceedings{yu2018spatio,
  title={Spatio-temporal graph convolutional networks: a deep learning framework for traffic forecasting},
  author={Yu, Bing and Yin, Haoteng and Zhu, Zhanxing},
  booktitle={Proceedings of the 27th International Joint Conference on Artificial Intelligence},
  pages={3634--3640},
  year={2018}
}

@inproceedings{wu2019graph,
  title={Graph WaveNet for Deep Spatial-Temporal Graph Modeling},
  author={Wu, Z and Pan, S and Long, G and Jiang, J and Zhang, C},
  booktitle={The 28th International Joint Conference on Artificial Intelligence (IJCAI)},
  year={2019},
  organization={International Joint Conferences on Artificial Intelligence Organization}
}

@article{bai2020adaptive,
  title={Adaptive graph convolutional recurrent network for traffic forecasting},
  author={Bai, Lei and Yao, Lina and Li, Can and Wang, Xianzhi and Wang, Can},
  journal={Advances in neural information processing systems},
  volume={33},
  pages={17804--17815},
  year={2020}
}

@inproceedings{shao2022pre,
  title={Pre-training enhanced spatial-temporal graph neural network for multivariate time series forecasting},
  author={Shao, Zezhi and Zhang, Zhao and Wang, Fei and Xu, Yongjun},
  booktitle={Proceedings of the 28th ACM SIGKDD conference on knowledge discovery and data mining},
  pages={1567--1577},
  year={2022}
}

@inproceedings{fang2021spatial,
  title={Spatial-temporal graph ode networks for traffic flow forecasting},
  author={Fang, Zheng and Long, Qingqing and Song, Guojie and Xie, Kunqing},
  booktitle={Proceedings of the 27th ACM SIGKDD conference on knowledge discovery \& data mining},
  pages={364--373},
  year={2021}
}

@inproceedings{shang2021discrete,
  title={Discrete Graph Structure Learning for Forecasting Multiple Time Series},
  author={Shang, Chao and Chen, Jie and Bi, Jinbo},
  booktitle={International Conference on Learning Representations},
  year={2021}
}

@inproceedings{shao2022spatial,
  title={Spatial-temporal identity: A simple yet effective baseline for multivariate time series forecasting},
  author={Shao, Zezhi and Zhang, Zhao and Wang, Fei and Wei, Wei and Xu, Yongjun},
  booktitle={Proceedings of the 31st ACM international conference on information \& knowledge management},
  pages={4454--4458},
  year={2022}
}

@inproceedings{jiang2023spatio,
  title={Spatio-temporal meta-graph learning for traffic forecasting},
  author={Jiang, Renhe and Wang, Zhaonan and Yong, Jiawei and Jeph, Puneet and Chen, Quanjun and Kobayashi, Yasumasa and Song, Xuan and Fukushima, Shintaro and Suzumura, Toyotaro},
  booktitle={Proceedings of the AAAI conference on artificial intelligence},
  volume={37},
  number={7},
  pages={8078--8086},
  year={2023}
}

@incollection{steinbach2009knn,
  title={kNN: k-nearest neighbors},
  author={Steinbach, Michael and Tan, Pang-Ning},
  booktitle={The top ten algorithms in data mining},
  pages={165--176},
  year={2009},
  publisher={Chapman and Hall/CRC}
}

@article{kleijnen2009kriging,
  title={Kriging metamodeling in simulation: A review},
  author={Kleijnen, Jack PC},
  journal={European journal of operational research},
  volume={192},
  number={3},
  pages={707--716},
  year={2009},
  publisher={Elsevier}
}

@inproceedings{li2025fine,
  title={Fine-Grained Traffic Inference from Road to Lane via Spatio-Temporal Graph Node Generation},
  author={Li, Shuhao and Yang, Weidong and Cui, Yue and Liu, Xiaoxing and Meng, Lingkai and Ma, Lipeng and Zhang, Fan},
  booktitle={Proceedings of the 31st ACM SIGKDD Conference on Knowledge Discovery and Data Mining V. 2},
  pages={1529--1540},
  year={2025}
}

@inproceedings{liu2023cross,
  title={Cross-city few-shot traffic forecasting via traffic pattern bank},
  author={Liu, Zhanyu and Zheng, Guanjie and Yu, Yanwei},
  booktitle={Proceedings of the 32nd ACM international conference on information and knowledge management},
  pages={1451--1460},
  year={2023}
}

@inproceedings{hu2024prompt,
  title={Prompt-based spatio-temporal graph transfer learning},
  author={Hu, Junfeng and Liu, Xu and Fan, Zhencheng and Yin, Yifang and Xiang, Shili and Ramasamy, Savitha and Zimmermann, Roger},
  booktitle={Proceedings of the 33rd ACM International Conference on Information and Knowledge Management},
  pages={890--899},
  year={2024}
}

@inproceedings{jin2023transferable,
  title={Transferable graph structure learning for graph-based traffic forecasting across cities},
  author={Jin, Yilun and Chen, Kai and Yang, Qiang},
  booktitle={Proceedings of the 29th ACM SIGKDD conference on knowledge discovery and data mining},
  pages={1032--1043},
  year={2023}
}

@article{amato2020novel,
  title={A novel framework for spatio-temporal prediction of environmental data using deep learning},
  author={Amato, Federico and Guignard, Fabian and Robert, Sylvain and Kanevski, Mikhail},
  journal={Scientific reports},
  volume={10},
  number={1},
  pages={22243},
  year={2020},
  publisher={Nature Publishing Group UK London}
}

@article{aryandoust2022enhanced,
  title={Enhanced spatio-temporal electric load forecasts using less data with active deep learning},
  author={Aryandoust, Arsam and Patt, Anthony and Pfenninger, Stefan},
  journal={Nature Machine Intelligence},
  volume={4},
  number={11},
  pages={977--991},
  year={2022},
  publisher={Nature Publishing Group UK London}
}

@inproceedings{li2024st,
  title={ST-ABC: Spatio-temporal attention-based convolutional network for multi-scale lane-level traffic prediction},
  author={Li, Shuhao and Cui, Yue and Li, Libin and Yang, Weidong and Zhang, Fan and Zhou, Xiaofang},
  booktitle={2024 IEEE 40th International Conference on Data Engineering (ICDE)},
  pages={1185--1198},
  year={2024},
  organization={IEEE}
}

@inproceedings{zhao2024just,
  title={A Just-In-Time Framework for Continuous Routing},
  author={Zhao, Jing and Li, Lei and Zhang, Mengxuan and Luo, Zihan and Zhao, Xi and Zhou, Xiaofang},
  booktitle={2024 IEEE 40th International Conference on Data Engineering (ICDE)},
  pages={4600--4613},
  year={2024},
  organization={IEEE}
}

@article{wang2024timemixer,
  title={Timemixer: Decomposable multiscale mixing for time series forecasting},
  author={Wang, Shiyu and Wu, Haixu and Shi, Xiaoming and Hu, Tengge and Luo, Huakun and Ma, Lintao and Zhang, James Y and Zhou, Jun},
  journal={arXiv preprint arXiv:2405.14616},
  year={2024}
}

@inproceedings{li2018diffusion,
  title={Diffusion Convolutional Recurrent Neural Network: Data-Driven Traffic Forecasting},
  author={Li, Yaguang and Yu, Rose and Shahabi, Cyrus and Liu, Yan},
  booktitle={International Conference on Learning Representations},
  year={2018}
}

@inproceedings{liu2023spatio,
  title={Spatio-temporal adaptive embedding makes vanilla transformer sota for traffic forecasting},
  author={Liu, Hangchen and Dong, Zheng and Jiang, Renhe and Deng, Jiewen and Deng, Jinliang and Chen, Quanjun and Song, Xuan},
  booktitle={Proceedings of the 32nd ACM international conference on information and knowledge management},
  pages={4125--4129},
  year={2023}
}

@inproceedings{jin2022selective,
  title={Selective cross-city transfer learning for traffic prediction via source city region re-weighting},
  author={Jin, Yilun and Chen, Kai and Yang, Qiang},
  booktitle={Proceedings of the 28th ACM SIGKDD conference on knowledge discovery and data mining},
  pages={731--741},
  year={2022}
}

@article{mo2022cross,
  title={Cross-city multi-granular adaptive transfer learning for traffic flow prediction},
  author={Mo, Jiqian and Gong, Zhiguo},
  journal={IEEE Transactions on Knowledge and Data Engineering},
  volume={35},
  number={11},
  pages={11246--11258},
  year={2022},
  publisher={IEEE}
}

@inproceedings{guo2019attention,
  title={Attention based spatial-temporal graph convolutional networks for traffic flow forecasting},
  author={Guo, Shengnan and Lin, Youfang and Feng, Ning and Song, Chao and Wan, Huaiyu},
  booktitle={Proceedings of the AAAI conference on artificial intelligence},
  volume={33},
  number={01},
  pages={922--929},
  year={2019}
}

@article{li2023dynamic,
  title={Dynamic graph convolutional recurrent network for traffic prediction: Benchmark and solution},
  author={Li, Fuxian and Feng, Jie and Yan, Huan and Jin, Guangyin and Yang, Fan and Sun, Funing and Jin, Depeng and Li, Yong},
  journal={ACM Transactions on Knowledge Discovery from Data},
  volume={17},
  number={1},
  pages={1--21},
  year={2023},
  publisher={ACM New York, NY}
}

@inproceedings{wu2021inductive,
  title={Inductive graph neural networks for spatiotemporal kriging},
  author={Wu, Yuankai and Zhuang, Dingyi and Labbe, Aurelie and Sun, Lijun},
  booktitle={Proceedings of the AAAI conference on artificial intelligence},
  volume={35},
  number={5},
  pages={4478--4485},
  year={2021}
}

@inproceedings{dong2024heterogeneity,
  title={Heterogeneity-informed meta-parameter learning for spatiotemporal time series forecasting},
  author={Dong, Zheng and Jiang, Renhe and Gao, Haotian and Liu, Hangchen and Deng, Jinliang and Wen, Qingsong and Song, Xuan},
  booktitle={Proceedings of the 30th ACM SIGKDD conference on knowledge discovery and data mining},
  pages={631--641},
  year={2024}
}

@article{guo2021learning,
  title={Learning dynamics and heterogeneity of spatial-temporal graph data for traffic forecasting},
  author={Guo, Shengnan and Lin, Youfang and Wan, Huaiyu and Li, Xiucheng and Cong, Gao},
  journal={IEEE Transactions on Knowledge and Data Engineering},
  volume={34},
  number={11},
  pages={5415--5428},
  year={2021},
  publisher={IEEE}
}

@article{marisca2024graph,
  title={Graph-based forecasting with missing data through spatiotemporal downsampling},
  author={Marisca, Ivan and Alippi, Cesare and Bianchi, Filippo Maria},
  journal={arXiv preprint arXiv:2402.10634},
  year={2024}
}

@inproceedings{wu2025spatio,
  title={Spatio-Temporal Forecasting under Open-World Missingness with Adaptive Mixture-of-Experts},
  author={Wu, Chenyu and Ma, Zhipeng and Zhang, Junbo and Ke, Songyu and Zheng, Yu},
  booktitle={Proceedings of the 34th ACM International Conference on Information and Knowledge Management},
  pages={3366--3375},
  year={2025}
}

@article{ke2025geomae,
  title={GeoMAE: Masking Representation Learning for Spatio-Temporal Graph Forecasting with Missing Values},
  author={Ke, Songyu and Wu, Chenyu and Liang, Yuxuan and Qin, Huiling and Zhang, Junbo and Zheng, Yu},
  journal={arXiv preprint arXiv:2508.14083},
  year={2025}
}

@inproceedings{yang2025graph,
  title={Graph Structure Learning for Spatial-Temporal Imputation: Adapting to Node and Feature Scales},
  author={Yang, Xinyu and Sun, Yu and Chen, Xinyang and Zhang, Ying and Yuan, Xiaojie},
  booktitle={Proceedings of the AAAI Conference on Artificial Intelligence},
  volume={39},
  number={1},
  pages={959--967},
  year={2025}
}

@article{mei2025deep,
  title={Deep networks as denoising algorithms: Sample-efficient learning of diffusion models in high-dimensional graphical models},
  author={Mei, Song and Wu, Yuchen},
  journal={IEEE Transactions on Information Theory},
  year={2025},
  publisher={IEEE}
}

@article{yun2025posterior,
  title={Posterior inference with diffusion models for high-dimensional black-box optimization},
  author={Yun, Taeyoung and Om, Kiyoung and Lee, Jaewoo and Yun, Sujin and Park, Jinkyoo},
  journal={arXiv preprint arXiv:2502.16824},
  year={2025}
}

@article{ma2025efficient,
  title={Efficient diffusion models: A comprehensive survey from principles to practices},
  author={Ma, Zhiyuan and Zhang, Yuzhu and Jia, Guoli and Zhao, Liangliang and Ma, Yichao and Ma, Mingjie and Liu, Gaofeng and Zhang, Kaiyan and Ding, Ning and Li, Jianjun and others},
  journal={IEEE Transactions on Pattern Analysis and Machine Intelligence},
  year={2025},
  publisher={IEEE}
}

@article{cao2024survey,
  title={A survey on generative diffusion models},
  author={Cao, Hanqun and Tan, Cheng and Gao, Zhangyang and Xu, Yilun and Chen, Guangyong and Heng, Pheng-Ann and Li, Stan Z},
  journal={IEEE transactions on knowledge and data engineering},
  volume={36},
  number={7},
  pages={2814--2830},
  year={2024},
  publisher={IEEE}
}

@inproceedings{liu2023pristi,
  title={Pristi: A conditional diffusion framework for spatiotemporal imputation},
  author={Liu, Mingzhe and Huang, Han and Feng, Hao and Sun, Leilei and Du, Bowen and Fu, Yanjie},
  booktitle={2023 IEEE 39th International Conference on Data Engineering (ICDE)},
  pages={1927--1939},
  year={2023},
  organization={IEEE}
}

@inproceedings{wen2023diffstg,
  title={Diffstg: Probabilistic spatio-temporal graph forecasting with denoising diffusion models},
  author={Wen, Haomin and Lin, Youfang and Xia, Yutong and Wan, Huaiyu and Wen, Qingsong and Zimmermann, Roger and Liang, Yuxuan},
  booktitle={Proceedings of the 31st ACM international conference on advances in geographic information systems},
  pages={1--12},
  year={2023}
}

@inproceedings{zang2020moflow,
  title={Moflow: an invertible flow model for generating molecular graphs},
  author={Zang, Chengxi and Wang, Fei},
  booktitle={Proceedings of the 26th ACM SIGKDD international conference on knowledge discovery \& data mining},
  pages={617--626},
  year={2020}
}

@article{madhawa2019graphnvp,
  title={GraphNVP: An Invertible Flow-Based Model for Generating Molecular Graphs},
  author={Madhawa, Kaushalya and Ishiguro, Katsuhiko and Nakago, Kosuke and Abe, Motoki},
  year={2019}
}

@inproceedings{jiniterative,
  title={Iterative Refinement Graph Neural Network for Antibody Sequence-Structure Co-design},
  author={Jin, Wengong and Wohlwend, Jeremy and Barzilay, Regina and Jaakkola, Tommi S},
  booktitle={International Conference on Learning Representations}
}

@article{luo2023stg4traffic,
  title={STG4Traffic: A Survey and Benchmark of Spatial-Temporal Graph Neural Networks for Traffic Prediction},
  author={Luo, Xunlian and Zhu, Chunjiang and Zhang, Detian and Li, Qing},
  journal={arXiv preprint arXiv:2307.00495},
  year={2023}
}

@article{boeing2017osmnx,
  title={OSMnx: New methods for acquiring, constructing, analyzing, and visualizing complex street networks},
  author={Boeing, Geoff},
  journal={Computers, environment and urban systems},
  volume={65},
  pages={126--139},
  year={2017},
  publisher={Elsevier}
}

@article{liu2023largest,
  title={Largest: A benchmark dataset for large-scale traffic forecasting},
  author={Liu, Xu and Xia, Yutong and Liang, Yuxuan and Hu, Junfeng and Wang, Yiwei and Bai, Lei and Huang, Chao and Liu, Zhenguang and Hooi, Bryan and Zimmermann, Roger},
  journal={Advances in Neural Information Processing Systems},
  volume={36},
  pages={75354--75371},
  year={2023}
}

@inproceedings{wang2020pm2,
  title={Pm2. 5-gnn: A domain knowledge enhanced graph neural network for pm2. 5 forecasting},
  author={Wang, Shuo and Li, Yanran and Zhang, Jiang and Meng, Qingye and Meng, Lingwei and Gao, Fei},
  booktitle={Proceedings of the 28th international conference on advances in geographic information systems},
  pages={163--166},
  year={2020}
}

@article{qiu2024tfb,
  title={TFB: Towards Comprehensive and Fair Benchmarking of Time Series Forecasting Methods},
  author={Qiu, Xiangfei and Hu, Jilin and Zhou, Lekui and Wu, Xingjian and Du, Junyang and Zhang, Buang and Guo, Chenjuan and Zhou, Aoying and Jensen, Christian S and Sheng, Zhenli and others},
  journal={Proceedings of the VLDB Endowment},
  volume={17},
  number={9},
  pages={2363--2377},
  year={2024},
  publisher={VLDB Endowment}
}

@inproceedings{lyu2025autostf,
  title={Autostf: Decoupled neural architecture search for cost-effective automated spatio-temporal forecasting},
  author={Lyu, Tengfei and Zhang, Weijia and Deng, Jinliang and Liu, Hao},
  booktitle={Proceedings of the 31st ACM SIGKDD Conference on Knowledge Discovery and Data Mining V. 1},
  pages={985--996},
  year={2025}
}

@inproceedings{gao2024spatial,
  title={Spatial-Temporal-Decoupled Masked Pre-training for Spatiotemporal Forecasting},
  author={Gao, Haotian and Jiang, Renhe and Dong, Zheng and Deng, Jinliang and Ma, Yuxin and Song, Xuan},
  booktitle={Proceedings of the Thirty-ThirdInternational Joint Conference on Artificial Intelligence},
  year={2024},
  organization={International Joint Conferences on Artificial Intelligence Organization}
}
% \newpage

\appendix

\section{Robustness to Degraded Semantics}
\label{app:semantic_degradation}

To evaluate the sensitivity of GenST to the quality of semantic descriptions, we conduct semantic degradation experiments on METR-LA, PeMS03, PeMS04, and PeMS08, with the unobserved-node ratio fixed at 20\%. We consider two types of semantic perturbations. The generic template replaces the descriptions of a subset of nodes with a generic template, simulating missing or coarse-grained semantic information. Shuffled descriptions randomly permute node descriptions, simulating incorrect semantic matching or noisy annotations. The degradation ratios are set to 20\%, 50\%, and 80\%.

\textbf{Table~\ref{tab:a_semantics}} shows that Full semantics consistently achieves the best performance, while the errors generally increase as the degradation ratio becomes larger. This indicates that semantic information is an important condition for GenST to achieve high-quality zero-shot prediction, as it provides effective functional priors and heterogeneity cues for unobserved nodes. For example, on METR-LA, the MAE of Full semantic is 5.57, while it increases to 7.24 and 7.72 under 80\% Generic template and 80\% Shuffled descriptions, respectively. On PeMS04, the corresponding MAE increases from 27.40 to 35.14 and 36.91. Further comparing the two perturbation types, Shuffled descriptions usually cause a larger performance drop than the Generic template, suggesting that incorrect node-level semantic matching is more harmful than the absence of fine-grained semantic details. In other words, coarse-grained semantics mainly weaken the conditional guidance, whereas incorrect semantics may inject misleading priors into the generation process. These results show that the semantic branch determines the performance upper bound of GenST, while coordinates, structural context, and the generative latent prior can provide compensation when semantics are incomplete. Therefore, GenST can substantially benefit from high-quality semantic descriptions, yet it does not suffer from performance collapse under partial semantic missingness or noisy annotations, demonstrating the robustness of its multi-modal conditioning design in realistic deployment scenarios.
\begin{table}[t]
  \centering
  \caption{Results of semantic degradation robustness.}
  \resizebox{\linewidth}{!}{
     \begin{tabular}{ccccccccc}
    \toprule
    \multirow{2}[4]{*}{\textbf{Dataset}} & \multirow{2}[4]{*}{\textbf{Metric}} & \textbf{Full} & \multicolumn{3}{c}{\textbf{Generic template}} & \multicolumn{3}{c}{\textbf{Shuffled descriptions}} \\
\cmidrule{4-9}          &       & \textbf{Semantic} & \textbf{20\%} & \textbf{50\%} & \textbf{80\%} & \textbf{20\%} & \textbf{50\%} & \textbf{80\%} \\
    \midrule
          & \textbf{MAE} & 5.57  & 6.13  & 6.61  & 7.24  & 6.38  & 6.92  & 7.72 \\
    \textbf{METR-LA} & \textbf{RMSE} & 9.89  & 10.76 & 11.5  & 12.58 & 11.08 & 11.93 & 13.25 \\
          & \textbf{MAPE} & 16.65\% & 18.28\% & 19.83\% & 21.74\% & 19.04\% & 20.66\% & 23.18\% \\
\cmidrule{2-9}          & \textbf{MAE} & 21.29 & 23.85 & 24.62 & 26.31 & 24.48 & 25.82 & 27.24 \\
    \textbf{PeMS03} & \textbf{RMSE} & 32.39 & 36.05 & 37.86 & 41.08 & 36.99 & 38.11 & 40.64 \\
          & \textbf{MAPE} & 40.74\% & 45.63\% & 49.30\% & 51.41\% & 47.07\% & 49.63\% & 53.13\% \\
\cmidrule{2-9}          & \textbf{MAE} & 27.4  & 30.33 & 32.18 & 35.14 & 31.07 & 33.56 & 36.91 \\
    \textbf{PeMS04} & \textbf{RMSE} & 41.41 & 45.92 & 48.61 & 52.97 & 46.85 & 50.72 & 55.31 \\
          & \textbf{MAPE} & 20.23\% & 22.34\% & 23.76\% & 26.11\% & 22.96\% & 24.82\% & 27.46\% \\
\cmidrule{2-9}          & \textbf{MAE} & 19.76 & 20.41 & 22.42 & 24.9  & 22.13 & 23.11 & 25.66 \\
    \textbf{PeMS08} & \textbf{RMSE} & 28.81 & 30.37 & 33.31 & 36.3  & 32.35 & 34.15 & 36.89 \\
          & \textbf{MAPE} & 21.40\% & 22.29\% & 23.22\% & 26.96\% & 24.00\% & 25.11\% & 27.91\% \\
    \bottomrule
    \end{tabular}%
    }
  \label{tab:a_semantics}%
\end{table}%

\begin{figure}[t]
% \vspace{-0.15cm}
% \hspace{0.9cm}
\centerline{\includegraphics[width=0.5\textwidth]{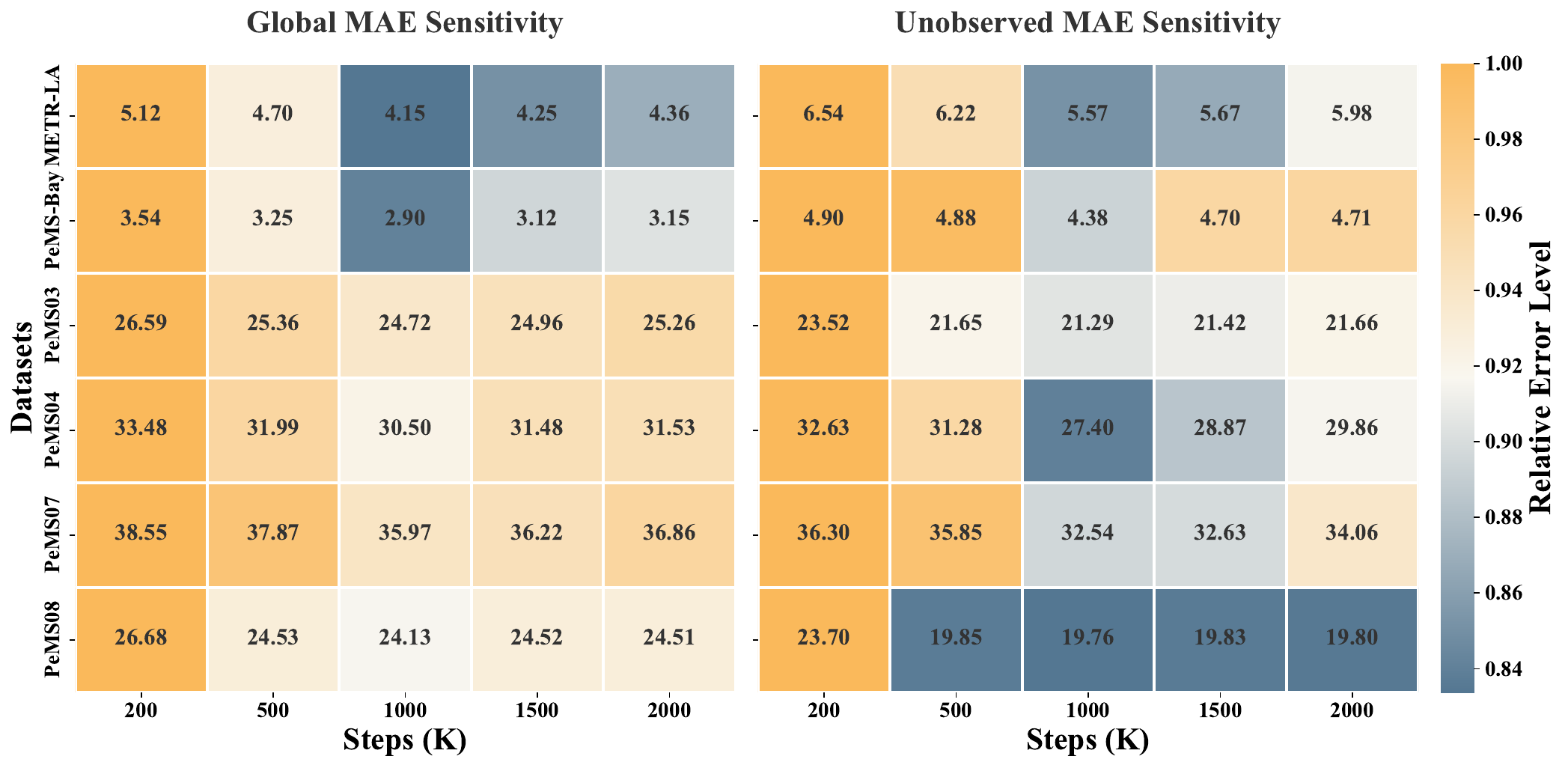}}
% \vspace{-0.3cm}
\caption{Performance sensitivity analysis across varying diffusion iteration counts.}
% \vspace{-0.4cm}
\label{fig: App_diff_step}
\end{figure}

\section{Diffusion Step $K$ Sensitivity Analysis \label{apsec:diff_step}}

To investigate the impact of diffusion steps $K$ on model performance, we conduct sensitivity experiments across all six datasets with $K \in \{200, 500, 1000, 1500, 2000\}$. \textbf{Figure~\ref{fig: App_diff_step}} presents the variation in Global and Unobserved MAE through heatmaps.

The results show that prediction accuracy follows a non-linear evolution, improving rapidly and then saturating as $K$ increases. Increasing $K$ from $200$ to $1000$ yields a significant reduction in MAE across all benchmarks. This suggests that low iteration counts prevent the Generative Transformer from fully learning the mapping from Gaussian noise to complex spatio-temporal dynamics, leading to coarse generated trajectories. GenST achieves the lowest mean MAE for both global and unobserved tasks at $K=1000$ on all six datasets, confirming the validity of this default parameter. Further increasing $K$ to $1500$ or $2000$ brings no additional gain. PeMS03 and PeMS08 stay close to a plateau, while METR-LA, PeMS-Bay, PeMS04, and PeMS07 show clear rebounds at $K=2000$, e.g., the Unobserved MAE on PeMS04 rises from $27.40$ to $29.86$. This decline likely arises from cumulative rounding errors or overfitting to latent noise details, which degrades zero-shot generalization. Thus, increasing iterations beyond a certain point yields no further accuracy gains while significantly increasing computational overhead. Furthermore, the unobserved task on PeMS04 and PeMS08 exhibits the highest sensitivity to $K$, where insufficient steps cause noticeable scale bias. For instance, on PeMS08 the Unobserved MAE of $23.70$ at $K=200$ is reduced to $19.76$ at $K=1000$, a relative reduction of 16.6\%, as sufficient iterations restore flow features.

\begin{table}[t]
  \centering
  \caption{Stability under different random seeds.}
  \resizebox{\linewidth}{!}{
    \begin{tabular}{ccccccc}
    \toprule
    \textbf{Dataset} & \textbf{Seed 1} & \textbf{Seed 2} & \textbf{Seed 3} & \textbf{Seed 4} & \textbf{Seed 5} & \textbf{Mean±Std} \\
    \midrule
    \textbf{METR-LA} & 5.57  & 5.42  & 5.71  & 5.63  & 5.49  & 5.56 ± 0.11 \\
    \textbf{PeMS04} & 27.4  & 27.12 & 27.78 & 27.95 & 27.31 & 27.51 ± 0.34 \\
    \textbf{PeMS03} & 21.29 & 20.94 & 21.66 & 21.48 & 21.12 & 21.30 ± 0.28 \\
    \textbf{PeMS08} & 19.76 & 19.42 & 20.08 & 19.89 & 19.60  & 19.75 ± 0.25 \\
    \bottomrule
    \end{tabular}%
    }
  \label{tab:app_seed}%
\end{table}%

\section{Stability under Different Random Seeds}
\label{app:random_seed_stability}

The main experimental results reported in the paper are averaged over three random seeds to reduce the influence of randomness. To further examine whether the reported performance depends on a particular random configuration, we conduct an additional five-seed stability study on different datasets, with the unobserved-node ratio fixed at 20\%. For each random seed, the unobserved-node split, model training process, and stochastic factors in generative inference are jointly varied. Therefore, this experiment provides a holistic stability check, evaluating whether GenST is sensitive to a specific node split, training initialization, or sampling noise.

As shown in \textbf{Table~\ref{tab:app_seed}}, GenST maintains stable performance across the four datasets. Specifically, the Unobserved MAE values on METR-LA, PeMS03, PeMS04, and PeMS08 are $5.56 \pm 0.11$, $21.30 \pm 0.28$, $27.51 \pm 0.34$, and $19.75 \pm 0.25$, respectively. The small standard deviations relative to the mean values indicate that the performance of GenST does not fluctuate substantially under different random seeds. This suggests that the main results are not caused by a favorable unobserved-node split, training initialization, or generative sampling noise, but instead reflect the stable inductive generalization ability of GenST under the FUNS setting.

\begin{figure}[t]
    \centering
    % \vspace{-0.3cm}
    \includegraphics[width=0.7\linewidth]{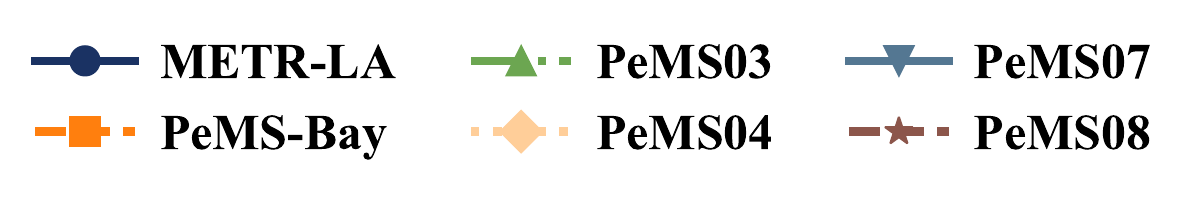}
    \vspace{0.1cm}
    \begin{subfigure}[b]{0.48\linewidth}
        \centering
        \includegraphics[width=\linewidth]{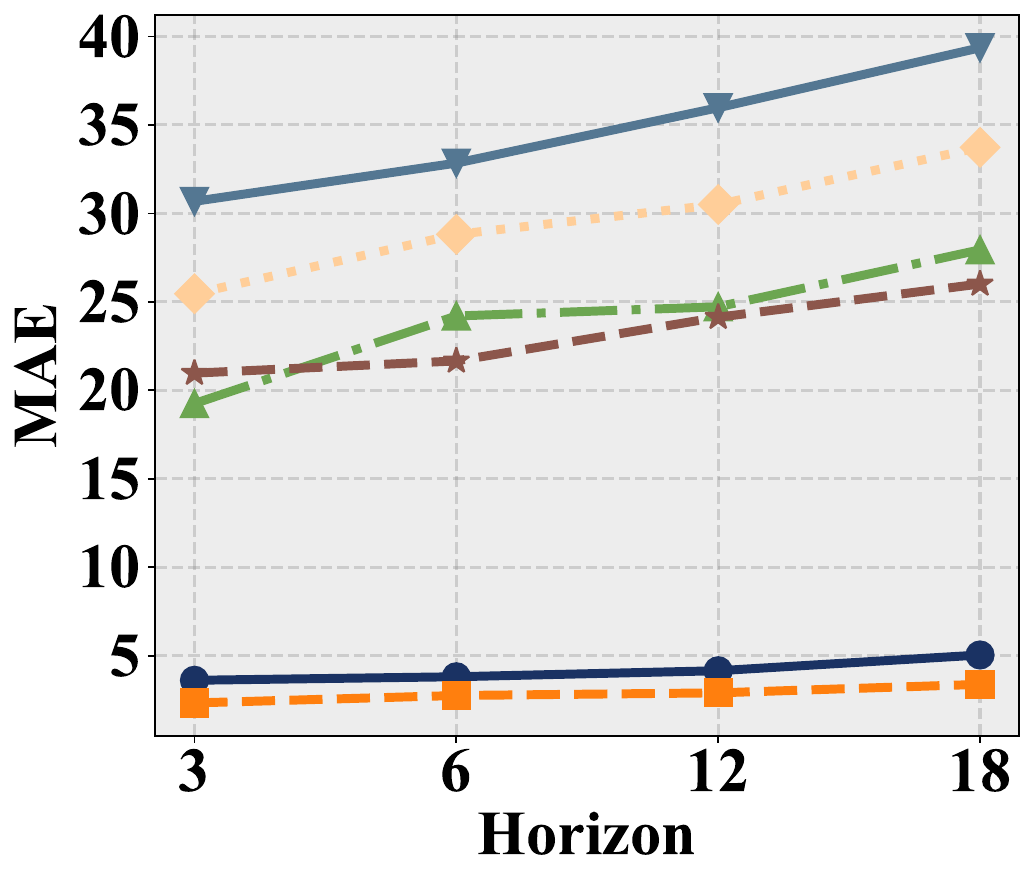}
        \caption{Global network results.}
    \end{subfigure}
    \hfill 
    \begin{subfigure}[b]{0.48\linewidth}
        \centering
        \includegraphics[width=\linewidth]{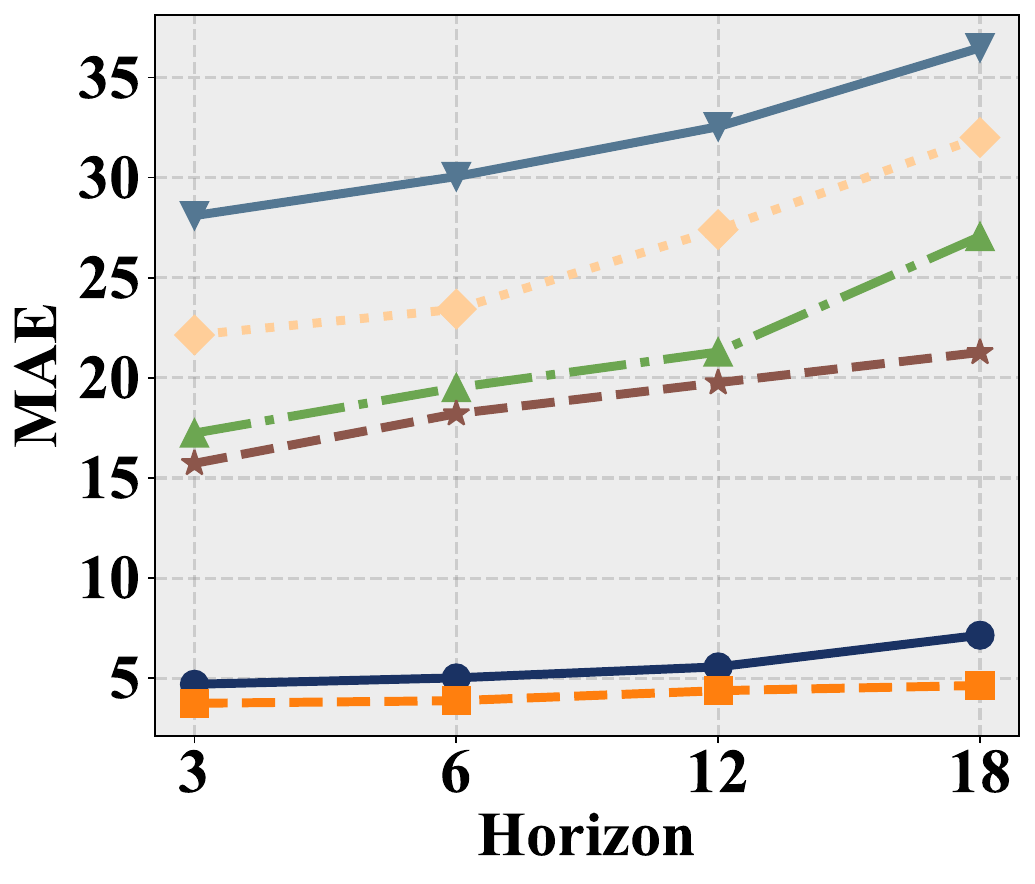}
        \caption{Unobserved nodes results.}
    \end{subfigure}
    \vspace{-0.2cm}
    \caption{Performance across varying forecasting horizons.}
    \label{fig:App_horizon}
    \vspace{-0.2cm}
\end{figure}

\section{Horizon Analysis \label{apsec:horizon}}
To further evaluate GenST across various prediction timescales, we examine the impact of prediction horizon $h \in \{3, 6, 12, 18\}$ while maintaining a constant historical window $T=12$.

As shown in \textbf{Figure~\ref{fig:App_horizon}}, prediction errors across all datasets exhibit a gradual upward trend as $h$ increases, consistent with the physical principle of temporal uncertainty accumulation in spatio-temporal forecasting. Notably, GenST maintains low MAE levels without prediction collapse even in the extreme setting of $h=18$, representing a 90-minute long-term prediction. For instance, on the challenging PeMS07 dataset, Unobserved MAE remains robustly around 35 as the horizon extends from 12 to 18 steps. The error growth slope in flow datasets is slightly steeper than in speed datasets, primarily because traffic flow experiences higher heterogeneity due to the increased impact of sporadic events and signal control over longer durations. Comparing figures (a) and (b), the error growth curves for the Unobserved regions are nearly parallel to those of the Global task. This consistency highlights a core advantage of GenST: the model learns universal, cross-node spatio-temporal evolution patterns rather than merely fitting historical numerical values from specific sensors. Even for nodes lacking any historical records, GenST utilizes trajectories generated from noise via the diffusion model to achieve long-term prediction robustness comparable to that of observed nodes.

\begin{figure}[t]
    \centering
    \includegraphics[width=0.99\linewidth]{figures/exp/missing/Legend_missing.pdf}
    \vspace{0.1cm}
    \begin{subfigure}[b]{0.48\linewidth}
        \centering
        \includegraphics[width=\linewidth]{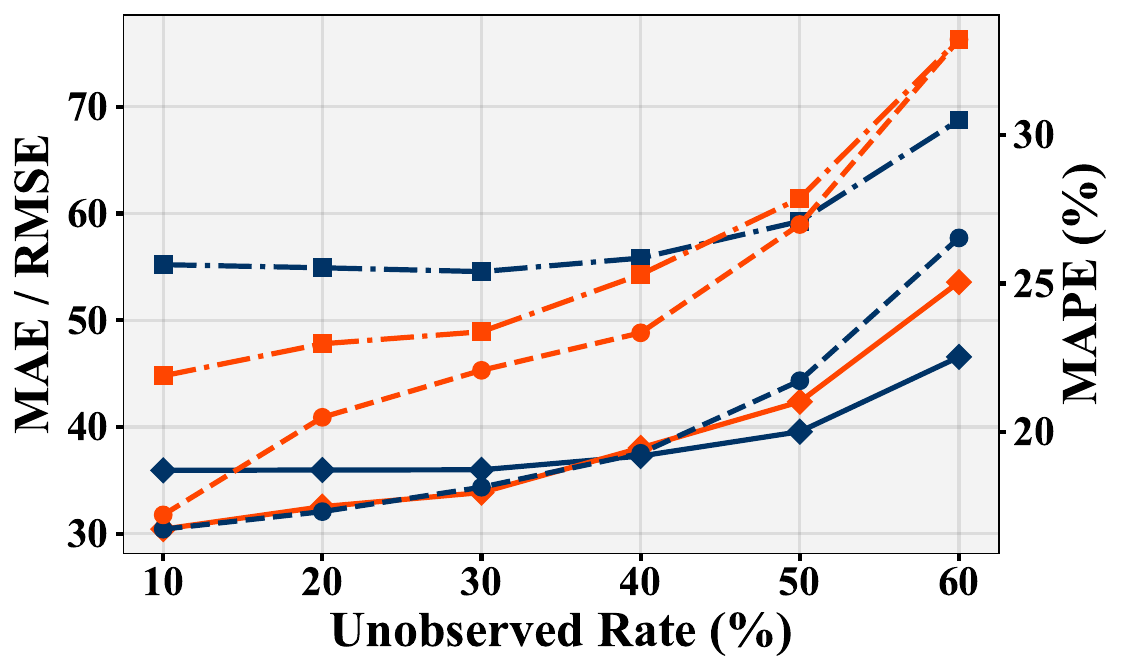}
        \caption{PeMS07}
    \end{subfigure}
    \hfill 
    \begin{subfigure}[b]{0.48\linewidth}
        \centering
        \includegraphics[width=\linewidth]{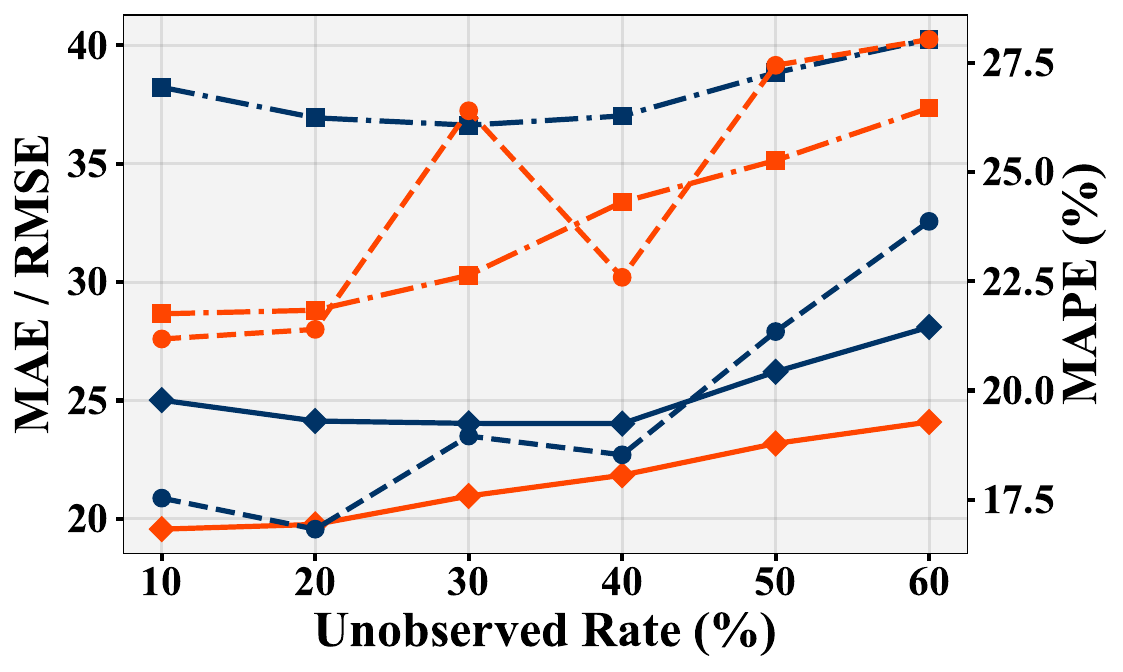}
        \caption{PeMS08}
    \end{subfigure}
    \caption{Supplemental robustness evaluation under varying unobserved node ratios.}
    \label{fig:App_missing}
\end{figure}
\section{Supplementary Experimental Analysis \label{apsec:supplementary}}
\subsection{Additional Unobserved Ratios Analysis}
\textbf{Figure~\ref{fig:App_missing}} presents performance curves for PeMS07 and PeMS08 with unobserved node ratios from 10\% to 60\%. Consistent with main experiments, GenST demonstrates robust stability across these large-scale networks. On the largest dataset, PeMS07, Unobserved MAE (orange line) exhibits only flat, linear growth rather than exponential divergence, confirming that the multi-modal semantic alignment module provides reliable priors even under extreme sparsity. Notably, PeMS08 (\textbf{Figure~\ref{fig:App_missing} (b)}) replicates the non-monotonic fluctuations observed in METR-LA, where errors at 30\%–40\% missingness slightly exceed those at 50\%. This recurrence strengthens our sample bias hypothesis: at lower ratios, random masking may disproportionately target hard nodes with irregular patterns, whereas higher ratios allow the unobserved set to better approximate the global distribution, stabilizing evaluation metrics.

\subsection{Additional Study of Backend LLMs}
\begin{figure}[t]
    \begin{subfigure}[b]{0.48\linewidth}
        \centering
        \includegraphics[width=\linewidth]{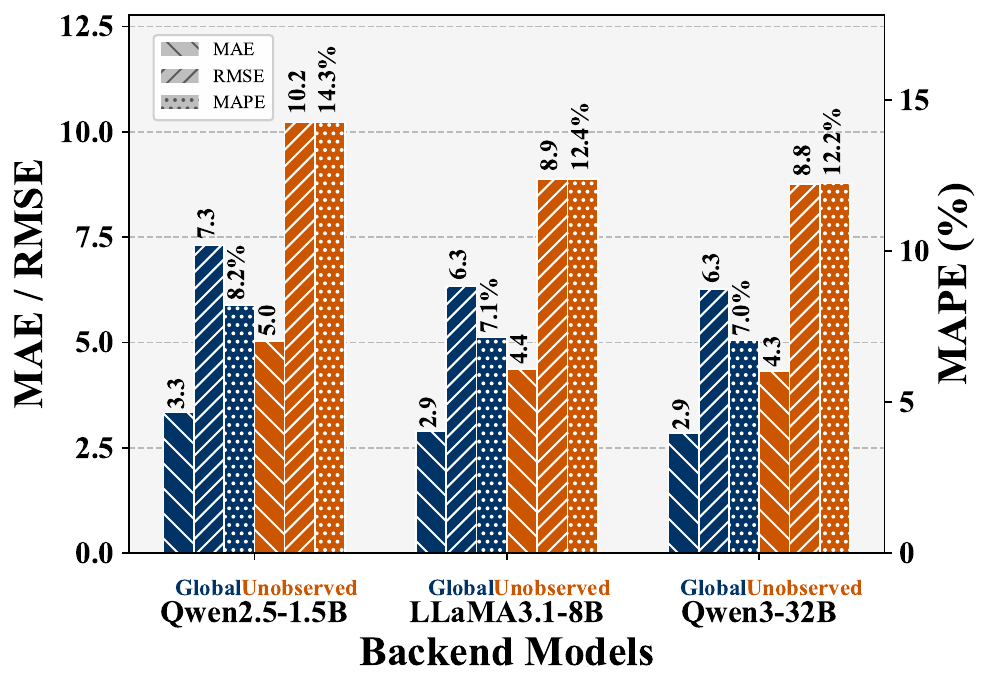}
        \caption{Accuracy on PeMS-Bay.}
    \end{subfigure}
    \hfill 
    \begin{subfigure}[b]{0.48\linewidth}
        \centering
        \includegraphics[width=\linewidth]{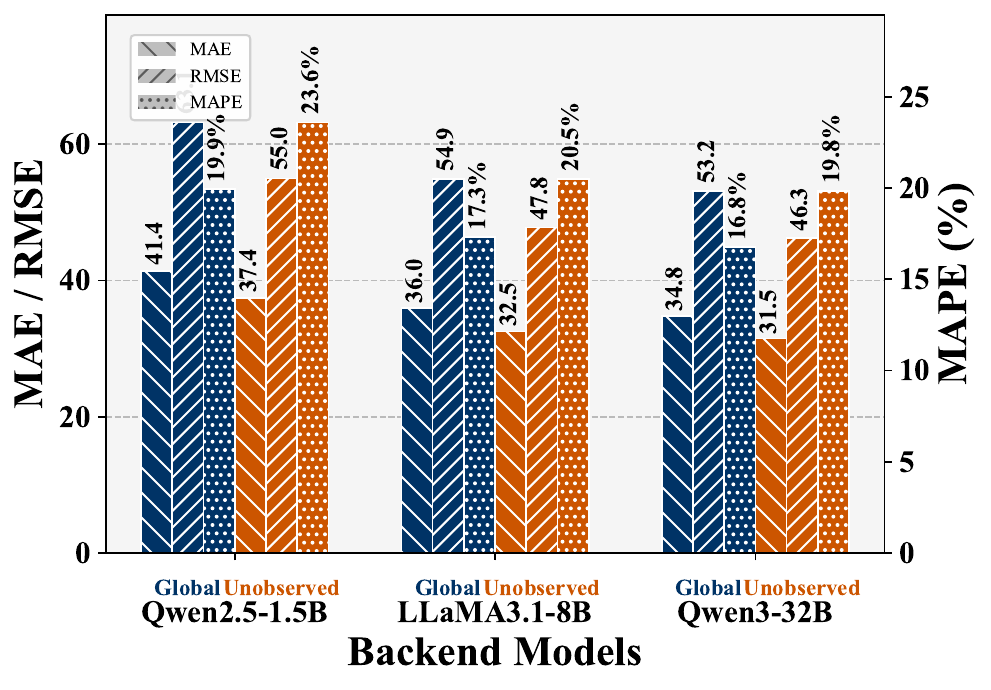}
        \caption{Accuracy on PeMS07.}
    \end{subfigure}
    \vspace{0.2cm} 
    \begin{subfigure}[b]{0.48\linewidth}
        \centering
        \includegraphics[width=\linewidth]{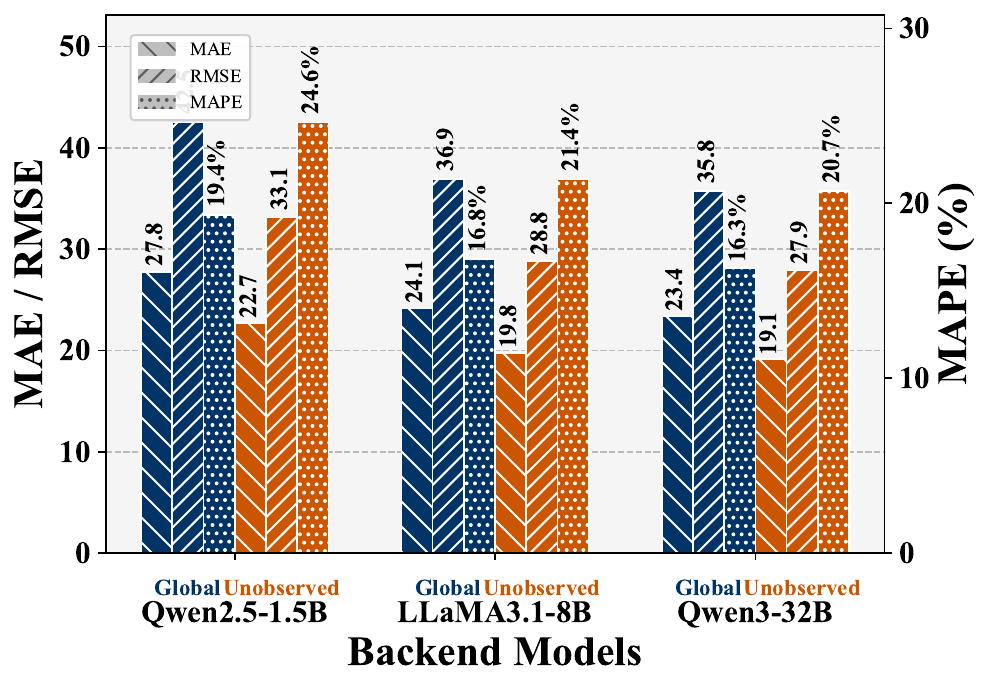}
        \caption{Accuracy on PeMS08.}
    \end{subfigure}
    \hfill 
    \begin{subfigure}[b]{0.48\linewidth}
        \centering
        \includegraphics[width=\linewidth]{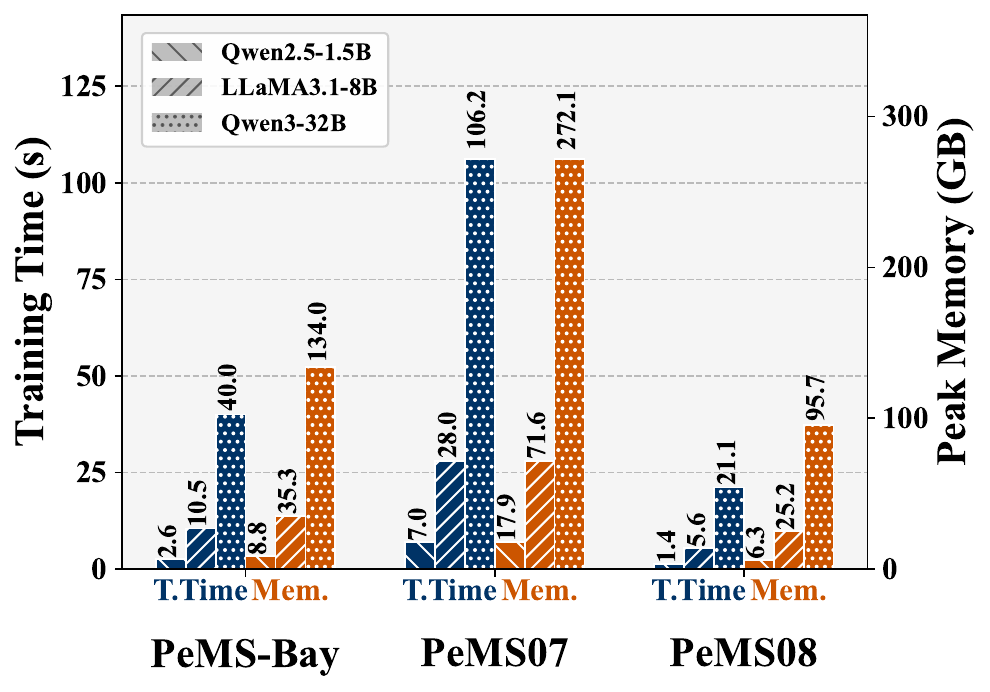}
        \caption{Training efficiency.}
    \end{subfigure}

    \caption{Supplemental backend LLM comparison across additional real-world traffic datasets.}
    \label{fig:App_backend}
\end{figure}

To verify the universality of LLM scaling effects, we extended the comparison of Qwen2.5-1.5B, Llama3.1-8B, and Qwen3-32B to PeMS-Bay, PeMS07, and PeMS08. Results  in \textbf{Figure~\ref{fig:App_backend}} remain highly consistent across all datasets: semantic reasoning capability positively correlates with zero-shot accuracy. Particularly in the complex PeMS07 flow network, the 32B model significantly outperforms the 1.5B version with a substantial MAE reduction. This indicates that more powerful LLMs better decode complex road semantics to infer physically grounded traffic magnitudes amidst high spatial heterogeneity. These expanded results also further highlight the computational bottlenecks encountered when processing large-scale graph data with massive backends.

\subsection{Ablation Study on Large-Scale Datasets}
\textbf{Table~\ref{tab:App_pems_abulation}} details ablation experiments on the largest datasets, PeMS07 and PeMS08, which largely align with our findings on smaller networks. The Semantic Alignment Module remains the most critical component; its removal from PeMS07 causes the sharpest performance drop, with Unobserved MAE surging by 11.29 and MAPE worsening by 17.74\%. On PeMS08, this variant results in a 13.96-point increase in MAE. There, the ST-VAE remains the second most vital module, and removing it increases Unobserved MAE by 5.57, a larger impact than removing Cross-Attention (+4.53) or Spatial Context (+3.16). On PeMS07, Cross-Attention ranks second (+2.09), while removing ST-VAE (+1.00) or Spatial Context (+0.82) causes smaller degradations comparable to seed variance. This suggests that on the largest and most heterogeneous network, conditional semantic injection contributes more than latent compression, whereas mapping sparse traffic states to compact latent manifolds remains important for suppressing noise on PeMS08.

\begin{table}[t]
  \centering
  \caption{PeMS07 and PeMS08 ablation results.}
  \resizebox{0.95\linewidth}{!}{
    \setlength{\tabcolsep}{1.5pt}
    \begin{tabular}{ccccc}
    \toprule
    \multirow{2}[3]{*}{\textbf{Variants}} & \textbf{Datasets} & \multicolumn{3}{c}{\textbf{PeMS07}} \\
    \cmidrule(lr){3-5}
          & \textbf{Metrics} & \textbf{MAE} & \textbf{RMSE} & \textbf{MAPE} \\
    \midrule
    \textbf{w/o Semantic} & \textbf{Global} & 40.05 \textcolor{myblue}{(\(\uparrow\)4.08)} & 63.54 \textcolor{myblue}{(\(\uparrow\)8.63)} & 20.10\% \textcolor{myblue}{(\(\uparrow\)2.77\%)} \\
    \textbf{Alignment} & \textbf{Unobserved} & 43.83 \textcolor{myorange}{(\(\uparrow\)11.29)} & 66.86 \textcolor{myorange}{(\(\uparrow\)19.07)} & 38.24\% \textcolor{myorange}{(\(\uparrow\)17.74\%)} \\
    \midrule
    \textbf{w/o} & \textbf{Global} & 37.39 \textcolor{myblue}{(\(\uparrow\)1.42)} & 59.61 \textcolor{myblue}{(\(\uparrow\)4.70)} & 18.49\% \textcolor{myblue}{(\(\uparrow\)1.16\%)} \\
    \textbf{ST-VAE} & \textbf{Unobserved} & 33.54 \textcolor{myorange}{(\(\uparrow\)1.00)} & 51.08 \textcolor{myorange}{(\(\uparrow\)3.29)} & 21.82\% \textcolor{myorange}{(\(\uparrow\)1.32\%)} \\
    \midrule
    \textbf{w/o Cross-} & \textbf{Global} & 38.10 \textcolor{myblue}{(\(\uparrow\)2.13)} & 59.60 \textcolor{myblue}{(\(\uparrow\)4.69)} & 17.99\% \textcolor{myblue}{(\(\uparrow\)0.66\%)} \\
    \textbf{Attention} & \textbf{Unobserved} & 34.63 \textcolor{myorange}{(\(\uparrow\)2.09)} & 50.48 \textcolor{myorange}{(\(\uparrow\)2.69)} & 21.66\% \textcolor{myorange}{(\(\uparrow\)1.16\%)} \\
    \midrule
    \textbf{w/o Spatial} & \textbf{Global} & 36.83 \textcolor{myblue}{(\(\uparrow\)0.86)} & 57.83 \textcolor{myblue}{(\(\uparrow\)2.92)} & 17.62\% \textcolor{myblue}{(\(\uparrow\)0.29\%)} \\
    \textbf{Context} & \textbf{Unobserved} & 33.36 \textcolor{myorange}{(\(\uparrow\)0.82)} & 48.64 \textcolor{myorange}{(\(\uparrow\)0.85)} & 20.83\% \textcolor{myorange}{(\(\uparrow\)0.33\%)} \\
    \midrule
    \midrule
    \multirow{2}[3]{*}{\textbf{Variants}} & \textbf{Datasets} & \multicolumn{3}{c}{\textbf{PeMS08}} \\
    \cmidrule(lr){3-5}
          & \textbf{Metrics} & \textbf{MAE} & \textbf{RMSE} & \textbf{MAPE} \\
    \midrule
    \textbf{w/o Semantic} & \textbf{Global} & 28.33 \textcolor{myblue}{(\(\uparrow\)4.20)} & 42.23 \textcolor{myblue}{(\(\uparrow\)5.29)} & 19.34\% \textcolor{myblue}{(\(\uparrow\)2.51\%)} \\
    \textbf{Alignment} & \textbf{Unobserved} & 33.72 \textcolor{myorange}{(\(\uparrow\)13.96)} & 47.75 \textcolor{myorange}{(\(\uparrow\)18.94)} & 36.40\% \textcolor{myorange}{(\(\uparrow\)15.00\%)} \\
    \midrule
    \textbf{w/o} & \textbf{Global} & 25.94 \textcolor{myblue}{(\(\uparrow\)1.81)} & 39.45 \textcolor{myblue}{(\(\uparrow\)2.51)} & 18.24\% \textcolor{myblue}{(\(\uparrow\)1.41\%)} \\
    \textbf{ST-VAE} & \textbf{Unobserved} & 25.33 \textcolor{myorange}{(\(\uparrow\)5.57)} & 33.69 \textcolor{myorange}{(\(\uparrow\)4.88)} & 23.21\% \textcolor{myorange}{(\(\uparrow\)1.81\%)} \\
    \midrule
    \textbf{w/o Cross-} & \textbf{Global} & 26.00 \textcolor{myblue}{(\(\uparrow\)1.87)} & 38.30 \textcolor{myblue}{(\(\uparrow\)1.36)} & 17.76\% \textcolor{myblue}{(\(\uparrow\)0.93\%)} \\
    \textbf{Attention} & \textbf{Unobserved} & 24.29 \textcolor{myorange}{(\(\uparrow\)4.53)} & 32.56 \textcolor{myorange}{(\(\uparrow\)3.75)} & 22.45\% \textcolor{myorange}{(\(\uparrow\)1.05\%)} \\
    \midrule
    \textbf{w/o Spatial} & \textbf{Global} & 25.20 \textcolor{myblue}{(\(\uparrow\)1.07)} & 36.98 \textcolor{myblue}{(\(\uparrow\)0.04)} & 17.05\% \textcolor{myblue}{(\(\uparrow\)0.22\%)} \\
    \textbf{Context} & \textbf{Unobserved} & 22.92 \textcolor{myorange}{(\(\uparrow\)3.16)} & 31.16 \textcolor{myorange}{(\(\uparrow\)2.35)} & 21.65\% \textcolor{myorange}{(\(\uparrow\)0.25\%)} \\
    \bottomrule
    \end{tabular}%
    }
  \label{tab:App_pems_abulation}
\end{table}

\begin{figure}[t]    
    \begin{subfigure}[b]{0.49\linewidth}
        \centering
        \includegraphics[width=\linewidth]{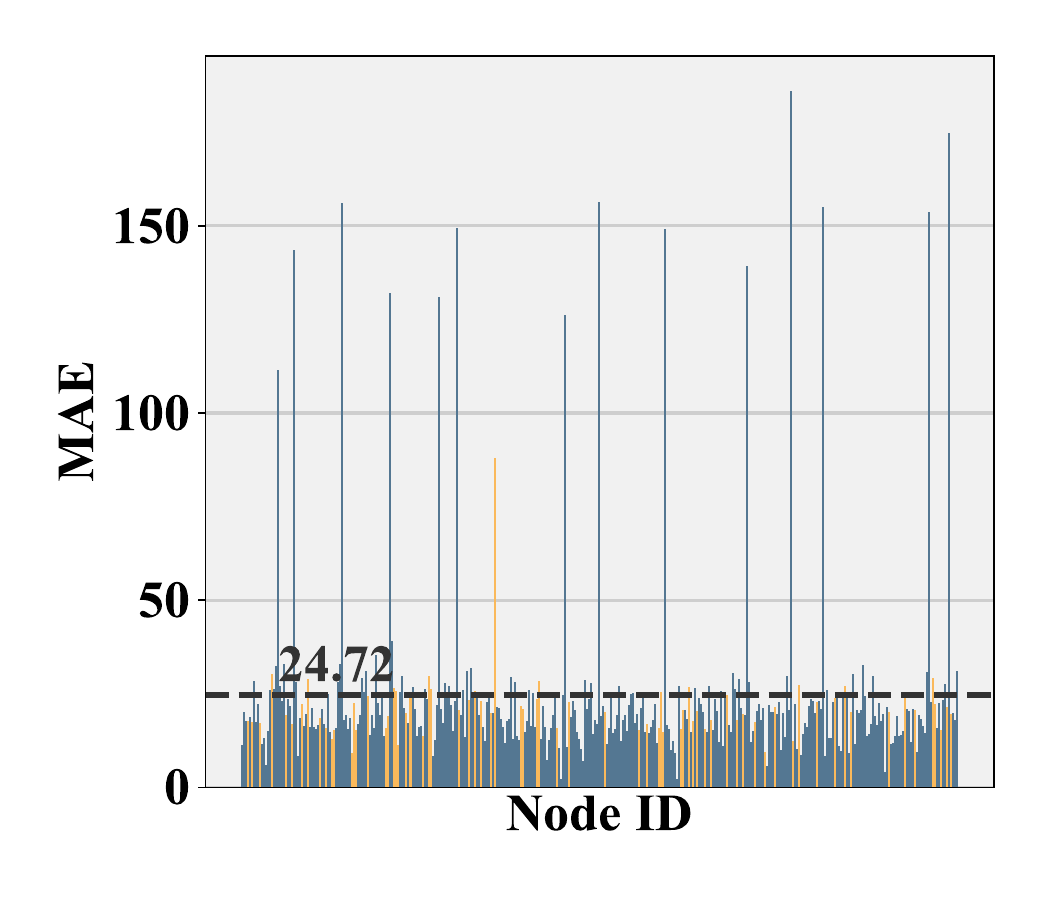}
        \caption{Global network view, observed nodes (blue).}
    \end{subfigure}
    \hfill 
    \begin{subfigure}[b]{0.49\linewidth}
        \centering
        \includegraphics[width=\linewidth]{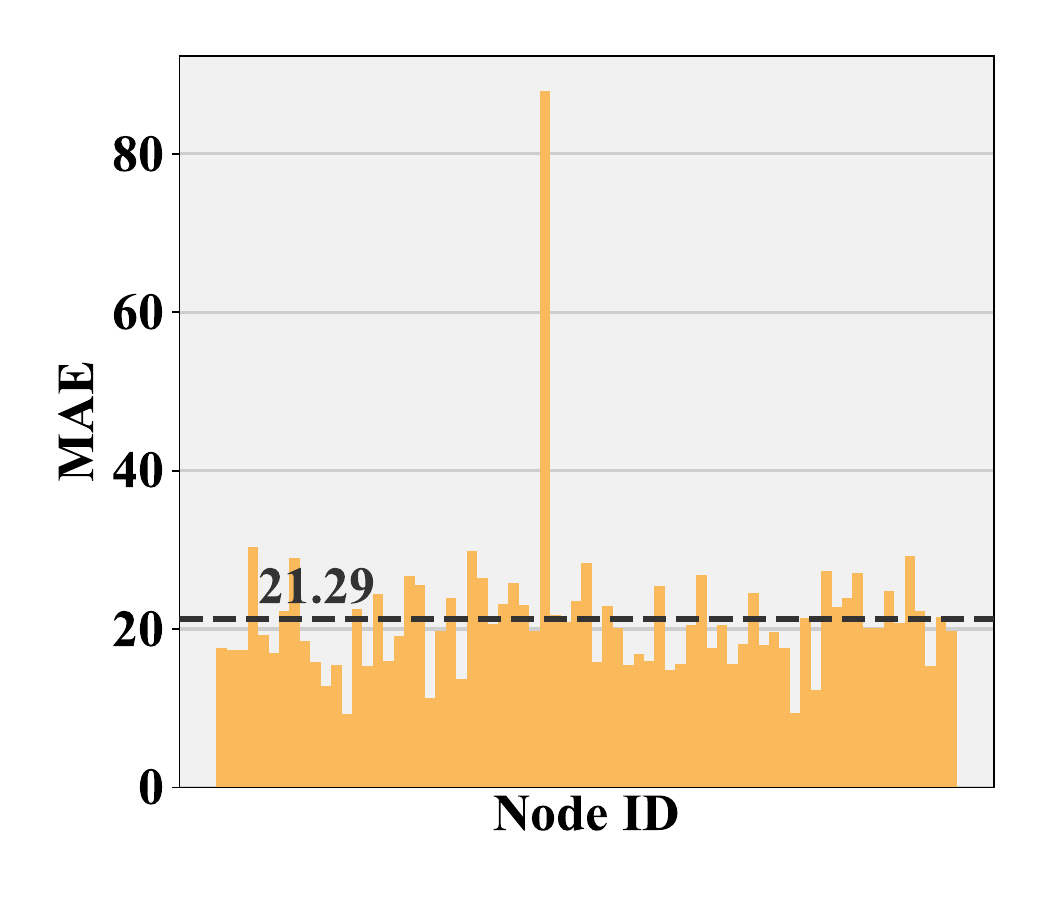}
        \caption{Unobserved nodes only (orange).}
    \end{subfigure}
    \caption{Per-node MAE distribution on PeMS03, dashed lines indicate subset means.}
    \label{fig:App_node_mae}
    \vspace{-0.3cm}
\end{figure}

\section{Dataset Statistics \label{Asec:Datasets}} 
We evaluate GenST on six traffic benchmark datasets and two non-traffic datasets for cross-domain validation. The traffic datasets include METR-LA and PeMS-Bay for traffic speed, and PeMS03, PeMS04, PeMS07, and PeMS08 for traffic flow~\cite{luo2023stg4traffic, li2023dynamic}, all sourced from the Caltrans Performance Measurement
System (PeMS) at a unified 5-minute resolution. For cross-domain generalization (\S\ref{sec:cross_domain}), we additionally adopt KnowAir~\cite{wang2020pm2} (184 air-quality monitoring stations with hourly PM2.5 concentrations) and Electricity~\cite{qiu2024tfb} (321 anonymized clients with hourly consumption records), both at 1-hour resolution. Following standard time-series benchmark conventions~\cite{qiu2024tfb}, all Electricity values are reported on the normalized scale, while traffic and air-quality datasets are reported in their native units. \textbf{Table~\ref{tab:datasets}} provides the comprehensive statistics for all eight datasets.

\begin{table}[t]
  \centering
  \caption{Statistics of all datasets.}
  \vspace{-0.3cm}
  \resizebox{\linewidth}{!}{
    \begin{tabular}{ccccc}
    \toprule
    Dataset       & METR-LA   & PeMS-Bay  & PeMS03     & PeMS04 \\
    \midrule
    Start Time    & 2012/3/1  & 2017/1/1  & 2018/9/1   & 2018/1/1 \\
    End Time      & 2012/6/30 & 2017/5/31 & 2018/11/30 & 2018/2/28 \\
    Time Interval & 5 min     & 5 min     & 5 min      & 5 min \\
    \# Object     & 207       & 325       & 358        & 307 \\
    \# Timestamps & 34272     & 52116     & 26208      & 16992 \\
    \midrule
    Dataset       & PeMS07    & PeMS08    & KnowAir    & Electricity \\
    \midrule
    Start Time    & 2017/5/1  & 2016/7/1  & 2016/1/1   & 2012/1/1 \\
    End Time      & 2017/8/31 & 2016/8/31 & 2023/12/31 & 2014/12/31 \\
    Time Interval & 5 min     & 5 min     & 1 hour     & 1 hour \\
    \# Object     & 883       & 170       & 184        & 321 \\
    \# Timestamps & 28224     & 17856     & 70128      & 26304 \\
    \bottomrule
    \end{tabular}%
    }
  \label{tab:datasets}%
\end{table}%

\subsection{Coordinate Acquisition.}
Sensor coordinates and adjacency matrices were collected from established open-source repositories: METR-LA and PeMS-Bay from the DCRNN repository~\cite{li2018diffusion}, PeMS03 from ASTGCN~\cite{guo2019attention}, and PeMS04/07/08 from the LargeST benchmark~\cite{liu2023largest}. All coordinates were linked via each dataset's canonical \texttt{sensor\_id} list as the primary key and cross-validated against the official Caltrans PeMS Station Inventory \footnote{https://pems.dot.ca.gov/} to ensure that longitude-latitude pairs reflect the true physical installation locations of sensors. For KnowAir we use the station coordinates and adjacency released by the PM2.5-GNN repository; Electricity is fully anonymized and contains no native coordinates, and we use the adjacency provided in its preprocessed release. Across all datasets, the unobserved node set $V^-$ is randomly drawn once per experimental run and held identical across the training, validation, and test partitions, ensuring a strict inductive zero-shot evaluation.

\subsection{Semantic Description Generation.} To construct textual descriptions for each sensor node (\textbf{Definition~4}), we designed a two-stage pipeline:

\noindent\textbf{(i) POI retrieval.} Using the OSMnx library~\cite{boeing2017osmnx}, we queried OpenStreetMap for Points of Interest (POIs) within a 500\,m radius of each sensor. The retrieved categories include: transportation facilities (\texttt{bus\_station}, \texttt{parking}, \texttt{fuel}, \texttt{motorway\_junction}, \texttt{toll\_gantry}), public services (\texttt{hospital}, \texttt{school}, \texttt{university}, \texttt{police}), commercial activity (\texttt{shop}, \texttt{office}, \texttt{tourism}), land use zones (\texttt{industrial}, \texttt{commercial}, \texttt{residential}), and rail transit (\texttt{railway station}, \texttt{subway\_entrance}).

\noindent\textbf{(ii) LLM-based description synthesis.} The retrieved POI profiles, together with sensor coordinates and road attributes, were fed to the DeepSeek API to generate a concise natural-language description characterizing each sensor's functional context. Two representative examples are shown below:

\begin{quote}
\small
\textbf{Sensor A} (commercial corridor): \emph{``This area is dominated by parking and fuel facilities, indicating high vehicle turnover and potential congestion from frequent stops. Minimal residential or office presence suggests traffic is primarily destination-based, with limited through-traffic.''}

\textbf{Sensor B} (industrial zone): \emph{``High industrial and office presence indicates strong commuter traffic, with highway nodes facilitating flow. Sensitive areas require speed management. Low commercial and residential activity suggests limited off-peak or pedestrian traffic.''}
\end{quote}

\noindent These descriptions serve as the input text, from which the LoRA-adapted LLM extracts semantic embeddings $\mathbf{s}_i$. By grounding each sensor in its surrounding urban context, this pipeline provides the semantic priors essential for zero-shot inference on unobserved nodes. All descriptions and coordinates are included in our released code repository.

\noindent\textbf{Extension to non-traffic datasets.} The same two-stage pipeline applies to KnowAir and Electricity by swapping only the structured inputs at stage (i). For KnowAir, each station's description is synthesized from its administrative city, a regional category drawn from a fixed five-label taxonomy (coastal, inland basin, industrial corridor, megacity, plateau), and aggregated meteorology. For Electricity, where no external metadata exists, descriptions are derived from training-period $V^+$ load statistics (mean daily consumption, peak hour, weekday-to-weekend ratio, summer-to-winter ratio, load-shape entropy) and translated by the LLM into client-type characterizations. In both cases, unobserved nodes $V^-$ receive a generic placeholder, and for Electricity, the Fourier coordinate branch $p_i$ is replaced by a shared learned $\texttt{<UNK\_LOC>}$ embedding. No load statistics or historical observations from unobserved nodes are used in this process.

\section{Baseline Adaptation \label{Asec: baseline}}
\subsection{Adaptation Strategies and Fairness Check}
Since most existing spatio-temporal forecasting baselines, particularly end-to-end deep learning models, are inherently transductive, they strictly rely on a node's own historical observations to infer future states. However, under the zero-shot setting of FUNS, history for unobserved nodes ($V^-$) is entirely missing, preventing direct inference. To evaluate these baselines fairly, we adopt a Neighborhood Aggregation-based Pseudo-History Construction strategy. Specifically, for each $u \in V^-$, we retrieve its first-order observed neighbors $\mathcal{N}(u) \cap V^+$ via graph topology and average their historical sequences to synthesize a "pseudo-history" as a proxy input. The input tensor $X_{t-T:t}^u$ is thus replaced by:
\begin{equation}
\hat{X}_{t-T:t}^u = \frac{1}{|\mathcal{N}(u)|} \sum_{v \in \mathcal{N}(u)} X_{t-T:t}^v.
\end{equation}

This adaptation enables autoregressive models like STGCN and GWNet to function by transforming the task into a neighborhood-mean-based regression. 

For isolated nodes without observed neighbors, we fill the gaps using the global average of all observed sensors. To examine whether the pseudo-history construction unfairly disadvantages stronger baselines, we further evaluate STEP and ST-LLM+ with their native missing-input strategies, namely mask token for STEP and ST-padding for ST-LLM+. On METR-LA, the relative difference between pseudo-history and native adaptation is within 1.3\% in terms of Unobserved MAE. This indicates that pseudo-history does not materially weaken these pretrained baselines. Rather, their main limitation under FUNS stems from the absence of node-specific historical observations, while GenST avoids this dependency through inductive multi-modal generation.

\subsection{Baseline Implementation Details}
We evaluated five categories of baselines with the following specific configurations:

\noindent\textbf{Statistical Learning}: Includes K-NN \cite{steinbach2009knn}, Kriging \cite{kleijnen2009kriging}, and HA \cite{campbell2008predicting}. Kriging and K-NN possess inherent spatial interpolation capabilities, leveraging spatial correlations directly to estimate unobserved values without requiring pseudo-history adaptation.

\noindent\textbf{End-to-End Deep Learning}: Includes STGCN \cite{yu2018spatio}, GWNet \cite{wu2019graph}, AGCRN \cite{bai2020adaptive}, STGODE \cite{fang2021spatial}, GTS \cite{shang2021discrete}, STID \cite{shao2022spatial}, and MegaCRN \cite{jiang2023spatio}. We strictly applied the pseudo-history construction strategy to these models. During training, these models essentially learn to map neighbor-averaged states to the target node's future state. While effective in spatially smooth speed fields, this simple averaging often over-smooths predictions in heterogeneous flow fields, failing to capture node-specific peak patterns—explaining their performance bottleneck in flow datasets.

\noindent\textbf{Pre-training \& Transfer Learning}: Includes STEP \cite{shao2022pre}, ST-LLM+ \cite{liu2025st}, TPB \cite{liu2023cross}, TransGTR \cite{jin2023transferable}, and STGP \cite{hu2024prompt}. Pre-trained models like STEP and ST-LLM+ rely on historical tokens as prompts; thus, we provide neighborhood-aggregated context tokens for zero-shot inference. For transfer models like TransGTR, we utilize their meta-learning or pattern-bank mechanisms. However, lacking a few-shot target samples for fine-tuning, these models must rely on universal parameters learned from the source domain.

\noindent\textbf{Graph Node Generation}: RoadDiff is currently the only generative baseline \cite{li2025fine}. Originally designed for road-to-lane inference, it relies on strict physical constraints (e.g., flow conservation and speed consistency). Since these hierarchical laws do not apply to the general topology of FUNS, we implemented a De-physicalization Adaptation. We retain the core Road-Lane Autoencoder and Lane Diffusion Module, mapping observed nodes as conditions and unobserved nodes as generation targets. Crucially, we removed all physical constraint losses ($\mathcal{L}_{con}$), allowing the diffusion backbone to learn latent nonlinear spatial correlations directly from the data distribution.

\section{Accuracy Metric Formulas \label{Asec:Metric}}
To rigorously assess GenST, we evaluate performance across two dimensions: Global (the entire network $V$) and Unobserved (the missing-history set $V^-$) over the forecasting horizon $h$.

\noindent\textbf{Mean Absolute Error (MAE)}: This metric quantifies the average linear magnitude of residuals. We calculate it for the global and unobserved sets as follows:
\begin{equation}
MAE_{global} = \frac{1}{h \cdot N}\sum^{h}_{t = 1}\sum_{r_i \in V} \lvert x^{r_i}_t - \hat{y}^{r_i}_t \rvert,
\end{equation}

\begin{equation}
MAE_{unobs} = \frac{1}{h \cdot N^-}\sum^{h}_{t = 1}\sum_{r_i \in V^-} \lvert x^{r_i}_t - \hat{y}^{r_i}_t \rvert.
\end{equation}

\noindent\textbf{Root Mean Squared Error (RMSE)}: RMSE highlights substantial deviations by applying a quadratic penalty to errors, reflecting the model's stability in handling peak fluctuations:
\begin{equation}RMSE_{global} = \sqrt{\frac{1}{h \cdot N}\sum^{h}_{t = 1}\sum_{r_i \in V} (x^{r_i}_t - \hat{y}^{r_i}_t)^2}, 
\end{equation}

\begin{equation}
    RMSE_{unobs} = \sqrt{\frac{1}{h \cdot N^-}\sum^{h}_{t = 1}\sum_{r_i \in V^-} (x^{r_i}_t - \hat{y}^{r_i}_t)^2}.
\end{equation}
\noindent\textbf{Mean Absolute Percentage Error (MAPE)}: MAPE evaluates the relative error by normalizing deviations against the ground truth, which is crucial for calibrating flow scales in zero-shot tasks:
\begin{equation}MAPE_{global} = \frac{1}{h \cdot N}\sum^{h}_{t = 1}\sum_{r_i \in V} \left\lvert \frac{x^{r_i}t - \hat{y}^{r_i}_t}{x^{r_i}_t} \right\rvert, 
\end{equation}

\begin{equation}
 MAPE_{unobs} = \frac{1}{h \cdot N^-}\sum^{h}_{t = 1}\sum_{r_i \in V^-} \left\lvert \frac{x^{r_i}_t - \hat{y}^{r_i}_t}{x^{r_i}_t} \right\rvert.
\end{equation}

\end{document}